\documentclass[letterpaper, 10 pt, conference]{ieeeconf}  % Comment this line out if you need a4paper

\IEEEoverridecommandlockouts                              % This command is only needed if 
\usepackage{graphicx}   % for pdf, bitmapped graphics files
\usepackage{epsfig} % for postscript graphics files
\usepackage{amsmath} % assumes amsmath package installed
\usepackage{amssymb}  % assumes amsmath package installed
\usepackage{caption}
\usepackage{subcaption}
\usepackage[hidelinks]{hyperref}

\usepackage{tikz}
\usepackage{mathdots}
\usepackage{yhmath}
\usepackage{cancel}
\usepackage{color}
\usepackage{siunitx}
\usepackage{array}
\usepackage{multirow}
\usepackage{amssymb}
\usepackage{gensymb}
\usepackage{tabularx}
\usepackage{extarrows}
\usepackage{booktabs}
\usetikzlibrary{fadings}
\usetikzlibrary{patterns}
\usetikzlibrary{shadows.blur}
\usetikzlibrary{shapes}
\usepackage{pgfplots}
\usepgfplotslibrary{fillbetween}

\usepackage{stfloats} % For floating figure*
\newcommand\norm[1]{\left\lVert#1\right\rVert}

\DeclareMathOperator*{\argmin}{arg\,min}

\newtheorem{rem}{Remark}

\usepackage{soul}
\usepackage{color}

\usetikzlibrary{external,calc,patterns,decorations.pathmorphing,decorations.markings,decorations.text,arrows,shapes,positioning, backgrounds}
\newcommand\copyrighttext{%
	\footnotesize \copyright 2022 IEEE.  Personal use of this material is permitted.  Permission from IEEE must be obtained for all other uses, in any current or future media, including reprinting/republishing this material for advertising or promotional purposes, creating new collective works, for resale or redistribution to servers or lists, or reuse of any copyrighted component of this work in other works.}
\newcommand\copyrightnotice{%
	\begin{tikzpicture}[remember picture,overlay]
		\node[anchor=south,yshift=10pt] at (current page.south) {\fbox{\parbox{\dimexpr\textwidth-\fboxsep-\fboxrule\relax}{\copyrighttext}}};
	\end{tikzpicture}%
}

\title{\LARGE \bf
Informed Sampling-based Collision Avoidance with Least Deviation from the Nominal Path
}

\author{Thomas T. Enevoldsen$^{1}$ and Roberto Galeazzi$^{1}$% <-this % stops a space
\thanks{This research is sponsored by the Danish Innovation Fund, The Danish Maritime Fund, Orients Fund and the Lauritzen Foundation through the Autonomy part of the ShippingLab project, Grant number 8090-00063B. The sea charts have been provided by the Danish Geodata Agency.}% <-this % stops a space
\thanks{$^{1}$Automation and Control Group, Department of Electrical and Photonics Engineering, Technical University of Denmark, DK-2800 Kgs. Lyngby, Denmark
        {\tt\small \{tthen,roga\}@dtu.dk}}%
}

\begin{document}

\maketitle
\thispagestyle{empty}
\pagestyle{empty}
\copyrightnotice

%%%%%%%%%%%%%%%%%%%%%%%%%%%%%%%%%%%%%%%%%%%%%%%%%%%%%%%%%%%%%%%%%%%%%%%%%%%%%%%%
\begin{abstract}
This paper addresses local path re-planning for $n$-dimensional systems by introducing an informed sampling scheme and cost function to achieve collision avoidance with minimum deviation from an (optimal) nominal path. The proposed informed subset consists of the union of ellipsoids along the specified nominal path, such that the subset efficiently encapsulates all points along the nominal path. The cost function penalizes large deviations from the nominal path, thereby ensuring current safety in the face of potential collisions while retaining most of the overall efficiency of the nominal path. The proposed method is demonstrated on scenarios related to the navigation of autonomous marine crafts.
\end{abstract}

%%%%%%%%%%%%%%%%%%%%%%%%%%%%%%%%%%%%%%%%%%%%%%%%%%%%%%%%%%%%%%%%%%%%%%%%%%%%%%%%
\section{INTRODUCTION}
The collision avoidance system is a crucial component for the safe motion control of autonomous systems seeking widespread adoption across different industrial sectors, such as collective mobility, precision farming, intermodal logistics, smart manufacturing. In all these industrial processes the operations carried out by or with the support of autonomous systems are characterized by some combination of metrics of efficiency -- e.g., minimum time, minimum energy, minimum distance --, and safety. The efficient execution of such operations implies the adherence to an (optimal) nominal path by the autonomous bus \cite{werling2010optimal,oliveira2019path}, autonomous ship \cite{Enevoldsen2022} or autonomous robot \cite{tsoularis1999avoiding}. At the mission planning stage it is hard to account for the dynamically changing local environments. Hence, there is the need for a path planner that computes optimal local deviations from the nominal path to ensure current safety while retaining most of the overall efficiency of nominal path, whenever a collision may disrupt the ongoing operation. 

Sampling-based motion planners are highly efficient at addressing complex planning problems with multiple constraints, such as those posed by collision avoidance and autonomous navigation tasks. Mechanisms and techniques for computing paths that minimize path length using sampling-based methods are widespread in current literature \cite{veras2019systematic,gammell2021asymptotically}, with one of the most influential methods being the Informed RRT* \cite{gammell2014informed}, which reduces the sampling space to an informed subset, contained within an ellipsoid, once an initial path is obtained. The informed set guarantees that it contains any point that can improve the solution, whilst increasing the probability that the solution cost is decreased by sampling within the subset. However, these methods typically seek to minimize path length, where for collision avoidance one may instead wish to efficiently compute paths with minimal deviation from a nominal. 

\begin{figure}[ptb]
     \centering
     \begin{subfigure}[b]{0.45\columnwidth}
         \centering
         \includegraphics[width=\textwidth,trim={2cm 1.5cm 2cm 1.8cm},clip]{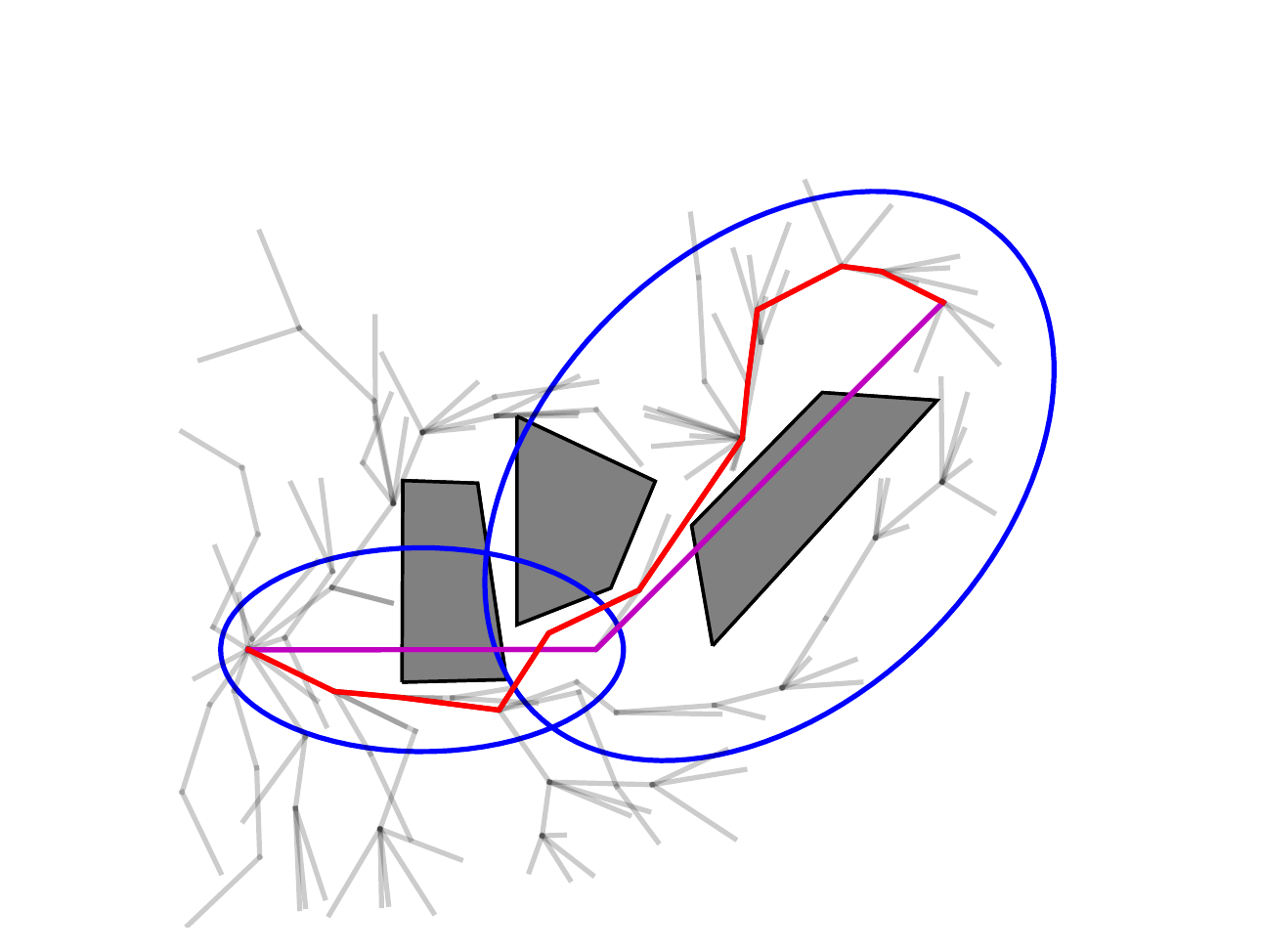}
         \caption{$N=250, \, c_d(\sigma) = 895$}
         \label{fig:teaser_n_250}
     \end{subfigure}
     \begin{subfigure}[b]{0.45\columnwidth}
         \centering
         \includegraphics[width=\textwidth,trim={2cm 1.5cm 2cm 1.8cm},clip]{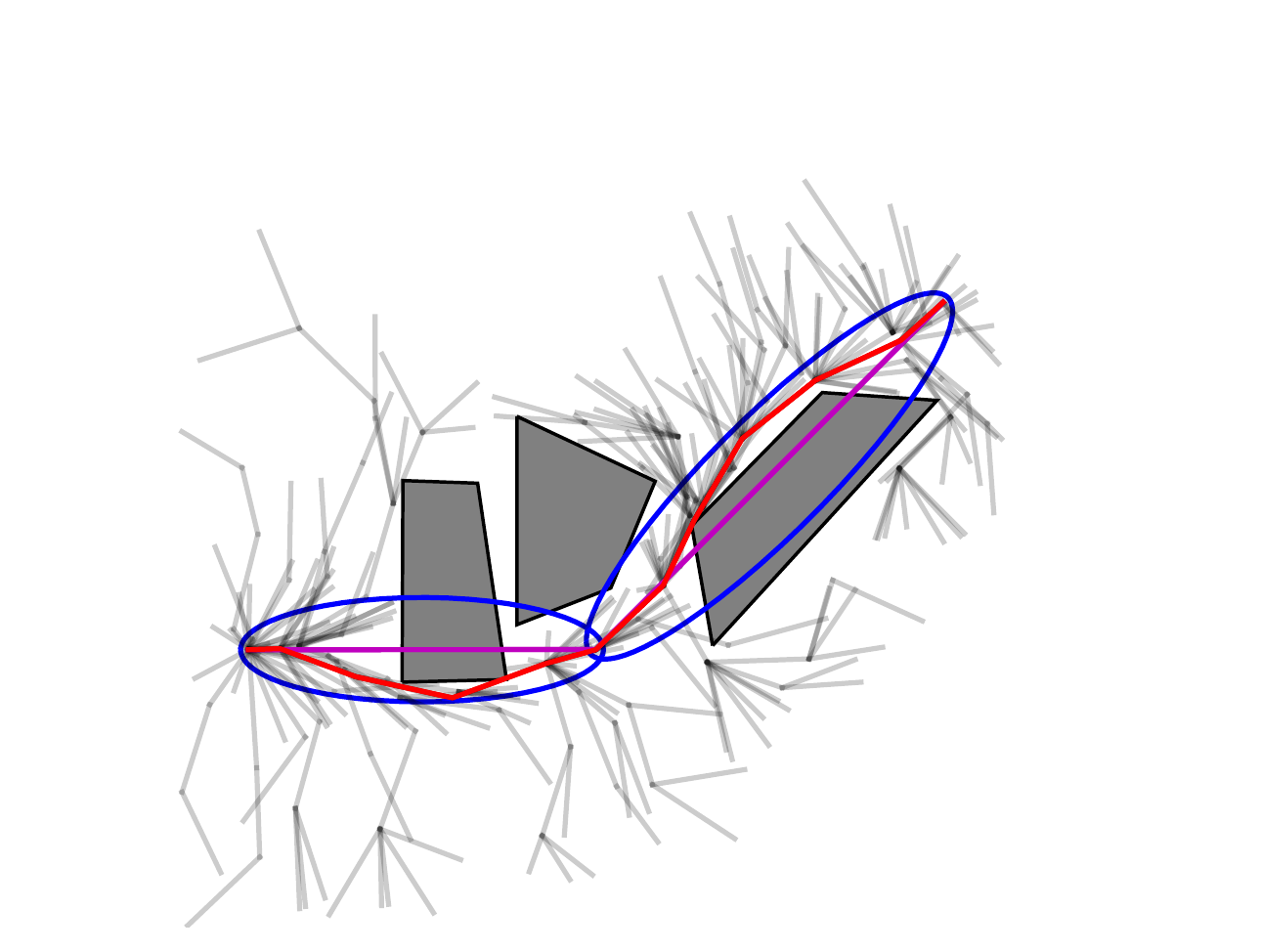}
         \caption{$N=500,\, c_d(\sigma) = 285$}
         \label{fig:teaser_n_500}
     \end{subfigure}
     \hfill
     \begin{subfigure}[b]{0.45\columnwidth}
         \centering
         \includegraphics[width=\textwidth,trim={2cm 1.5cm 2cm 1.8cm},clip]{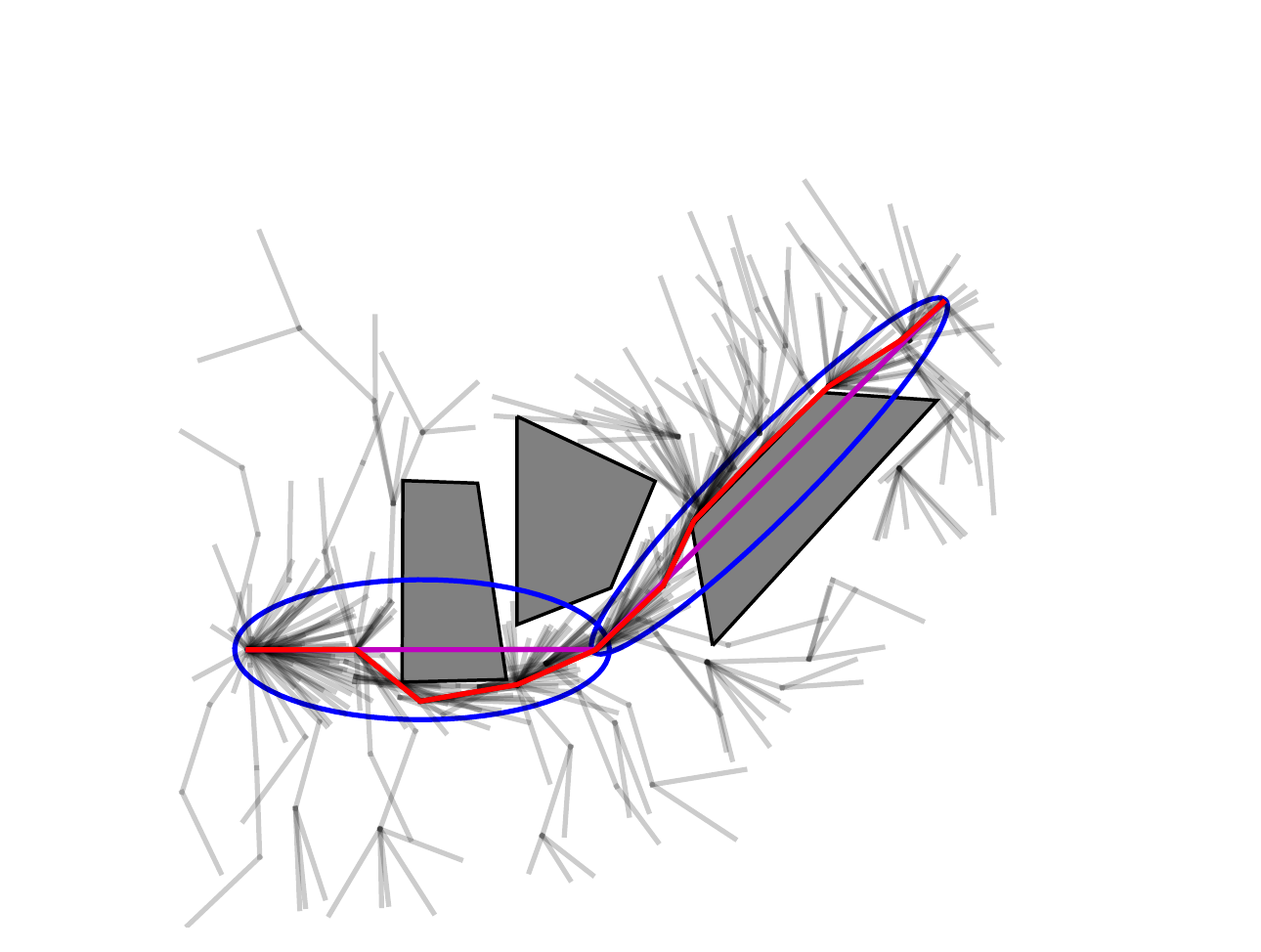}
         \caption{$N=750,\, c_d(\sigma) = 229$}
         \label{fig:teaser_n_750}
     \end{subfigure}
     \begin{subfigure}[b]{0.45\columnwidth}
         \centering
         \includegraphics[width=\textwidth,trim={2cm 1.5cm 2cm 1.8cm},clip]{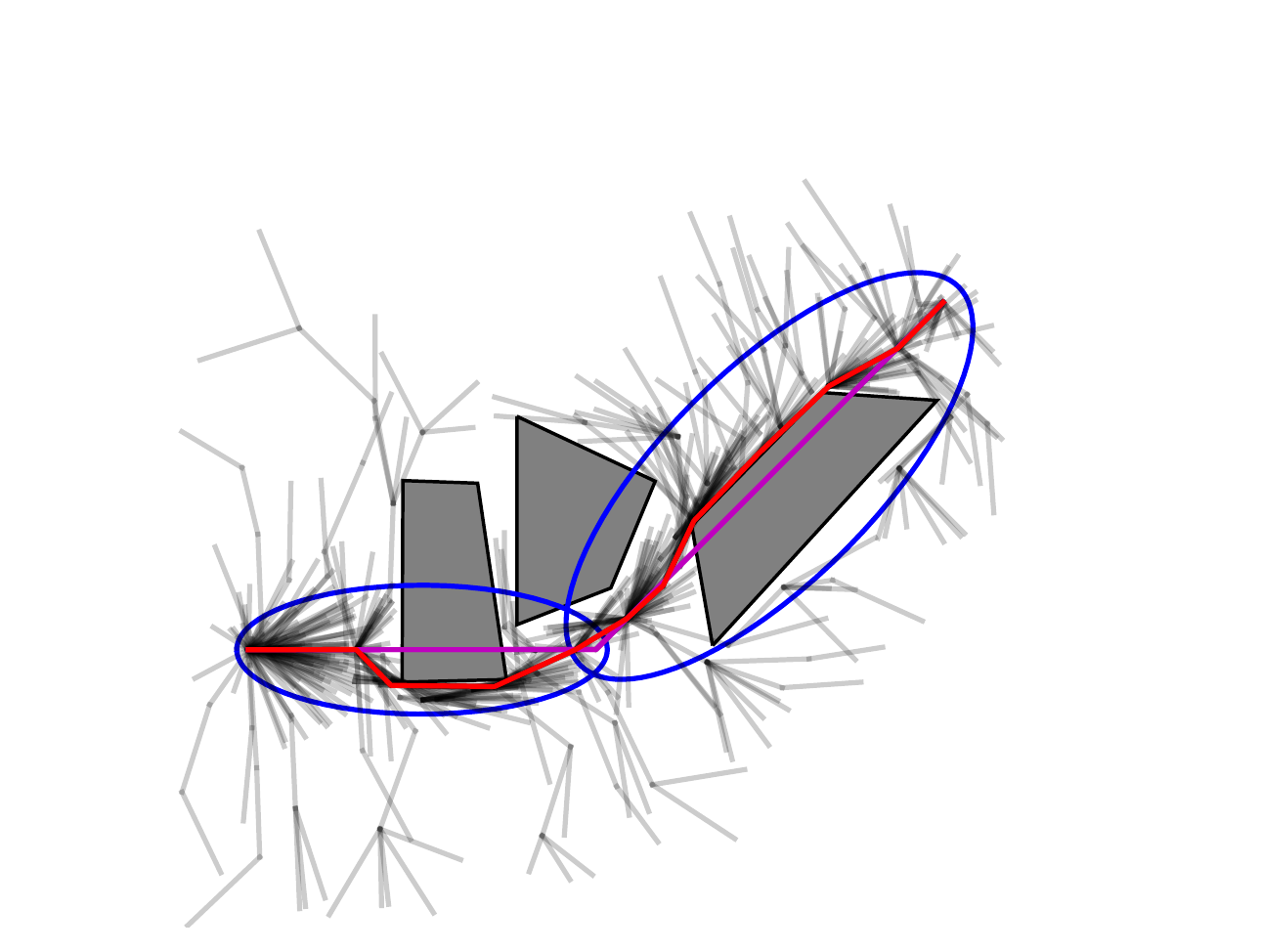}
         \caption{$N=1000,\, c_d(\sigma) = 206$}
         \label{fig:teaser_n_1000}
     \end{subfigure}
        \caption{Informed sampling strategy for computing the minimum path deviation, where an ellipsoidal subset is formed along each segment of the nominal trajectory. 
        The nominal trajectory (magenta) is obstructed by some obstacles, and therefore the planned path (red) is computed to circumvent these whilst minimizing the deviation from the nominal path.}
        \label{fig:teaser_figure}
        \vspace{-0.5cm}
\end{figure}

This paper focuses on local path re-planning for $n$-dimensional systems which have an (optimal) nominal path to achieve collision avoidance in dynamic environments, and it makes the following contributions. First, we propose an extension to the concept of the informed subset to allow for convergence towards solutions with minimum path deviation. This is achieved by introducing a cost function, which allows the underlying algorithm to minimize with respect to the nominal path. The extension involves forming multiple overlapping informed subsets along the nominal path, which results in an informed set composed of the union of multiple ellipsoidal subsets (Fig.~\ref{fig:teaser_figure}). Last, a switching condition and additional sampling biasing are proposed to allow for rapid convergence towards the nominal path.
% Teaser picture
% trim={<left> <lower> <right> <upper>}

\subsection{Related work}
Optimal Sampling-based Motion Planning (SBMP) came to fruition when \cite{karaman2011sampling} introduced RRT* and PRM*, which are asymptotically optimal in probability~\cite{gammell2021asymptotically}. 
To improve the convergence rate and performance of RRT*, \cite{gammell2014informed,gammell2018informed} proposed the Informed RRT*, which reduces the sampling space to an ellipsoidal subset once an initial solution is found. This increases the probability that each subsequent sample has a greater likelihood of improving the current best found solution. The concept of the informed subset then became an integral part of other SBMP algorithms  \cite{gammell2015batch,strub2020adaptively,choudhury2016regionally}.

Current research aims at extending the capabilities of the informed subset to further accelerate convergence to certain classes of solutions. Recently, \cite{mandalika2021guided} identifies smaller informed sets within the informed set itself, using the notion of a beacon, and thereby honing the search. In \cite{joshi2019non} the authors also propose a method for identifying subregions within the informed set.
\cite{enevoldsen2021colregs} specializes an informed sampling scheme that decreases the size of the search space to produce paths that abide by the maritime rules-of-the-road. \cite{li2021sliding} slides an informed subset along the found path, computing local solutions of minimum path length. The work by \cite{ryu2019improved} utilizes a pre-computed gridmap in order to find an initial solution quickly, such that the informed subset \cite{gammell2014informed} can be applied sooner.
\cite{yi2018generalizing} proposes a scheme for sampling generalized informed sets using Markov Chain Monte Carlo, allowing for arbitrarily shaped non-convex informed sets. \cite{sustarevas2021task} proposes to inform the planning algorithm about the manipulation task of mobile manipulators, i.e. a sequence of poses the end-effector must reach, by introducing it as success criterion when computing the movement of the mobile base.
\cite{gammell2021asymptotically} details a general overview of the state-of-the-art in optimal SBMP.

Within the fields of self-driving cars and autonomous marine systems, SBMP has gotten a foothold. \cite{paden2016survey} and  \cite{claussmann2019review} survey the application of various motion planning techniques for autonomous vehicles, providing insight into the use of SBMP for driving in urban environments and highways, respectively. \cite{lin2021sampling} proposes a method for repairing existing trajectories, where infeasible parts of the nominal trajectory are repaired to compute feasible deviations. \cite{tang2020reference} discretizes the nominal trajectory and places guide points in locations that are infeasible with the nominal, and thereby biases the sampling. \cite{vonasek2009rrt} uses an estimated nominal path, such as from a Voronoi graph, to guide the RRT exploration through cluttered environments. \cite{lan2015continuous} explores a sampling-based scheme that compute paths, which are similar in curvature to the nominal path.
Whereas within the realm of discrete planning, algorithms such as lifelong planning A* \cite{koenig2004lifelong} and D*-lite \cite{koenig2002d} concern efficient re-planning.

In the maritime domain, SBMP algorithms are favoured due to the existence of both the complex constraints and environments. \cite{chiang2018colreg} investigates using a non-holonomic RRT for collision avoidance.  \cite{Enevoldsen2022}, \cite{zaccone2020collision} and \cite{enevoldsen2021grounding} utilize RRT* for collision avoidance, taking various other metrics into account, such as minimizing nominal path deviation, speed loss, curvature and grounding risk. 

The reviewed literature emphasizes two main aspects: The informed set is a powerful and effective concept to channel the sampling effort of SBMP algorithms and achieve faster convergence to the optimal path; SBMP algorithms have been used to plan between the start and goal states for designing both nominal paths and path alterations along a single straight segment of a nominal path. This paper advances the application of informed SBMP algorithms to collision avoidance along multi-segment paths by introducing an extended informed set and a cost function that penalizes deviations from the nominal path.

\section{Preliminaries}
The general formulation of the optimal sampling-based motion planning problem is now presented, as well as the formalization of the informed subset as proposed by~\cite{gammell2014informed}.
\subsection{Optimal sampling-based motion planning}
Let $\mathcal{X} \subseteq \mathbb{R}^{n}$ be the state space, with $\mathbf{x}$ denoting the state. The state space is composed of two subsets: the free space $X_{\text{free}}$, and the obstacles $X_{\text {obs}}$, where $X_{\text{free}}=\mathcal{X} \backslash X_{\text{obs}}$. 
The states contained within $X_{\text{free}}$ are all states that are feasible with respect to the constraints posed by the system and the environment.
Let $\mathbf{x}_{\text{start}} \in X_{\text{free}}$ be the initial state at some time $t=0$ and $\mathbf{x}_{\text{end}} \in X_{\text{free}}$ the desired final state at some time $t=T$. 
Let $\sigma:[0,1] \mapsto X_{\text{free}}$ be a sequence of states that constitutes a found path, and $\Sigma$ be the set of all feasible and nontrivial paths. 
The objective is then to find the optimal path $\sigma^{*}$, which minimizes a cost function $c(\cdot)$, while connecting $\mathbf{x}_{\text{start}}$ to $\mathbf{x}_{\text{end}}$ through states $\mathbf{x}_i \in X_{\text{free}}$,
\begin{equation}
\begin{split}
\sigma^{*}=\underset{\sigma \in \Sigma}{\arg \min }\left\{c(\sigma) \mid \right. \sigma(0)=\mathbf{x}_{\text {start }},\, &\sigma(1)=\mathbf{x}_{\text {end }}, \\
\forall s \in[0,1],\, &\sigma(s) \in \left. X_{\text {free }}\right\}.
\end{split}
\end{equation}

The most commonly adopted cost function is the Euclidean path length, which gives rise to  the shortest path problem. Given a path $\sigma$ consisting of $n$ states, the Euclidean path length is given by
\begin{equation}
    c_l(\sigma) = \sum^{n}_{i=1} \norm{\mathbf{x}_i - \mathbf{x}_{i-1}}_2, \quad \forall \, \mathbf{x}_i \in \sigma .
\end{equation}
The cost function is additive, i.e. given a sequence of $n$ states and some index $k$ the following equality holds true
\begin{equation}
c \left( (\mathbf{x}_0,\dots,\mathbf{x}_n) \right) = c \left((\mathbf{x}_0,\dots,\mathbf{x}_k) \right) + c \left((\mathbf{x}_k,\dots,\mathbf{x}_n) \right)
\end{equation}
Therefore, whenever a new node or edge is added, the cost to go from the root to the nearest node, together with the cost from the nearest node to the new node, is computed as
\begin{equation}
    c(\sigma) = c \left((\mathbf{x}_{\text{start}},\dots,\mathbf{x}_{\text{nearest}})\right) + c \left((\mathbf{x}_{\text{nearest}},\mathbf{x}_{\text{new}}) \right) 
\end{equation}
as required by the underlying SBMP \cite{karaman2011sampling}.
%----- Informed Sampling

\subsection{Informed sampling}
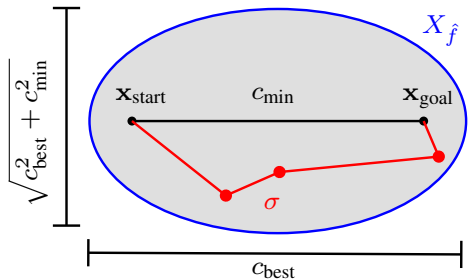
\begin{figure}[t]
    \centering
    \resizebox{0.75\columnwidth}{!}{\tikzset{every picture/.style={line width=0.75pt}} %set default line width to 0.75pt        

\begin{tikzpicture}[x=0.75pt,y=0.75pt,yscale=-1,xscale=1]
%uncomment if require: \path (0,993); %set diagram left start at 0, and has height of 993

%Shape: Ellipse [id:dp8474491510221505] 
\draw  [blue,fill=lightgray!50] (165.04,600.08) .. controls (165.04,574.67) and (199.62,554.06) .. (242.29,554.06) .. controls (284.96,554.06) and (319.55,574.67) .. (319.55,600.08) .. controls (319.55,625.5) and (284.96,646.1) .. (242.29,646.1) .. controls (199.62,646.1) and (165.04,625.5) .. (165.04,600.08) -- cycle ;
%Straight Lines [id:da7684044218827588] 
\draw    (182.46,600.08) -- (302.13,600.08) ;
\draw [shift={(302.13,600.08)}, rotate = 0] [color={rgb, 255:red, 0; green, 0; blue, 0 }  ][fill={rgb, 255:red, 0; green, 0; blue, 0 }  ][line width=0.75]      (0, 0) circle [x radius= 1.34, y radius= 1.34]   ;
\draw [shift={(182.46,600.08)}, rotate = 0] [color={rgb, 255:red, 0; green, 0; blue, 0 }  ][fill={rgb, 255:red, 0; green, 0; blue, 0 }  ][line width=0.75]      (0, 0) circle [x radius= 1.34, y radius= 1.34]   ;
%Straight Lines [id:da6955510155978231] 
\draw [color={rgb, 255:red, 255; green, 0; blue, 0 }  ,draw opacity=1 ]   (182.46,600.08) -- (221,630.67) ;
\draw [shift={(221,630.67)}, rotate = 38.43] [color={rgb, 255:red, 255; green, 0; blue, 0 }  ,draw opacity=1 ][fill={rgb, 255:red, 255; green, 0; blue, 0 }  ,fill opacity=1 ][line width=0.75]      (0, 0) circle [x radius= 2.01, y radius= 2.01]   ;
%Straight Lines [id:da500149871452646] 
\draw [color={rgb, 255:red, 255; green, 0; blue, 0 }  ,draw opacity=1 ]   (243,621) -- (308.33,614.67) ;
\draw [shift={(308.33,614.67)}, rotate = 354.46] [color={rgb, 255:red, 255; green, 0; blue, 0 }  ,draw opacity=1 ][fill={rgb, 255:red, 255; green, 0; blue, 0 }  ,fill opacity=1 ][line width=0.75]      (0, 0) circle [x radius= 2.01, y radius= 2.01]   ;
%Straight Lines [id:da5567980764602527] 
\draw [color={rgb, 255:red, 255; green, 0; blue, 0 }  ,draw opacity=1 ]   (302.13,600.08) -- (308.33,614.67) ;
%Straight Lines [id:da5070675720891411] 
\draw [color={rgb, 255:red, 255; green, 0; blue, 0 }  ,draw opacity=1 ]   (221,630.67) -- (243,621) ;
\draw [shift={(243,621)}, rotate = 336.28] [color={rgb, 255:red, 255; green, 0; blue, 0 }  ,draw opacity=1 ][fill={rgb, 255:red, 255; green, 0; blue, 0 }  ,fill opacity=1 ][line width=0.75]      (0, 0) circle [x radius= 2.01, y radius= 2.01]   ;
%Straight Lines [id:da49545186085850057] 
\draw    (155.61,643) -- (155.61,553.67) ;
\draw [shift={(155.61,553.67)}, rotate = 90] [color={rgb, 255:red, 0; green, 0; blue, 0 }  ][line width=0.75]    (0,5.59) -- (0,-5.59)   ;
\draw [shift={(155.61,643)}, rotate = 90] [color={rgb, 255:red, 0; green, 0; blue, 0 }  ][line width=0.75]    (0,5.59) -- (0,-5.59)   ;
%Straight Lines [id:da7855158103869768] 
\draw    (164.56,654.14) -- (317.56,654.83) ;
\draw [shift={(317.56,654.83)}, rotate = 180.26] [color={rgb, 255:red, 0; green, 0; blue, 0 }  ][line width=0.75]    (0,5.59) -- (0,-5.59)   ;
\draw [shift={(164.56,654.14)}, rotate = 180.26] [color={rgb, 255:red, 0; green, 0; blue, 0 }  ][line width=0.75]    (0,5.59) -- (0,-5.59)   ;

% Text Node
\draw (130,637) node [anchor=north west][inner sep=0.75pt]  [font=\footnotesize,rotate=-270] [align=left] {$\displaystyle \sqrt{c_{\text{best}}^{2} +c_{\text{min}}^{2}}$};
% Text Node
\draw (230,656.64) node [anchor=north west][inner sep=0.75pt]  [font=\footnotesize] [align=left] {$\displaystyle c_{\text{best}}$};
% Text Node
\draw (174.2,585) node [anchor=north west][inner sep=0.75pt]  [font=\footnotesize] [align=left] {$\mathbf{x}_{\text{start}}$};
% Text Node
\draw (292,585) node [anchor=north west][inner sep=0.75pt]  [font=\footnotesize] [align=left] {$\mathbf{x}_{\text{goal}}$};
% Text Node
\draw (230,585) node [anchor=north west][inner sep=0.75pt]  [font=\footnotesize] [align=left] {$\displaystyle c_{\text{min}}$};

% Text Node
\draw (300,555) node [text=blue][anchor=north west][inner sep=0.75pt]  [font=\footnotesize] [align=left] {$\displaystyle X_{\hat{f}}$};
% Text Node
\draw (235,630) node [anchor=north west][inner sep=0.75pt]  [font=\footnotesize, red] [align=left] {$\displaystyle \sigma$};

\end{tikzpicture}}
    \caption{The informed subset as proposed by \cite{gammell2014informed}, the sampling region is reduced to an ellipsoid, and thereby increasing the probability that the sampled states improve the found path.}
    \label{fig:standard_informedset}
    \vspace{-0.5cm}
\end{figure}

The concept of an informed sampling space was introduced by \cite{gammell2014informed} with the Informed RRT*. It was shown that the reduction of the sampling region to an informed subset increased the probability that each subsequent sample would improve the current best found solution. In the case of \cite{gammell2014informed,gammell2018informed} an informed subset for the euclidean distance was formulated as an ellipsoid, and was given by
\begin{equation}
    X_{\hat{f}} = \{ \mathbf{x} \in \mathcal{X} \, | \norm{\mathbf{x}_{\text{start}} - \mathbf{x}}_2 +  \norm{\mathbf{x} - \mathbf{x}_{\text{end}}}_2 \leq c_{\text{best}} \}
\end{equation}
with $\mathbf{x}_{\text{start}}$ and $\mathbf{x}_{\text{end}}$ representing the start and end states of a given path, and $c_{\text{best}}$ the path length of the current best found path. Once an initial path is obtained, one can form an informed subset that is scaled based on the the minimum possible path $c_{\text{min}}$ and $c_{\text{best}}$, as shown in Fig.~\ref{fig:standard_informedset}. The informed subset, which is a prolate hyperspheroid, represents all possible points that can improve the current solution cost, and allows one to sample these particular points directly. Generating samples within the ellipsoid can be done analytically, as described in \cite{gammell2014informed}.

Typically, the initial sampling scheme consists of uniformly sampling the state space, which is commonly achieved by uniformly sampling a $n$-dimensional hyperrectangle $\mathbf{x}_{\text{rand}} \sim \mathcal{U}(X_{\text{rect}})$. In order to ensure that it is favourable to switch to a given informed set, its Lebesgue measure is typically compared to that of the original sampling space
\begin{equation}\label{eq:nom_switching}
    \lambda \left(X_{\hat{f}}\right) < \lambda(X_{\text{rect}}) 
\end{equation}
where the Lebesgue measure of the ellipsoid is given by \cite{gammell2018informed}
\begin{equation}
    \lambda \left(X_{\hat{f}}\right) = \frac{c_{\text{best}} ( c_{\text{best}}^2 - c_{\text{min}}^2 )^{\frac{n - 1}{2}}}{2^n} \frac{\pi^{\frac{n}{2}}}{\Gamma\left(\frac{n}{2}+1\right)}
\end{equation}
with $c_{\text{best}}$ and $c_{\text{min}}$ as shown in Fig.~\ref{fig:standard_informedset}.
Further details regarding the informed subset can be found in \cite{gammell2014informed} and \cite{gammell2018informed}. 

\section{Informed Sampling for Collision Avoidance with Least Path Deviation}
The novel contribution of the paper is now introduced by formalizing the cost function for computing paths with minimum deviation, the associated informed space, a proposed sampling bias and the switching condition.

\subsection{Cost function for minimum path deviation}\label{sec:deviation_cost}
Let $\sigma^{\text{nom}}$ be the nominal path, i.e. the sequence of $m$ states $\mathbf{x}^{\text{nom}}_i \in \mathcal{X}$ that connect $\mathbf{x}_{\text{start}}$ and $\mathbf{x}_{\text{end}}$. It is assumed that two consecutive states, $\mathbf{x}^{\text{nom}}_i$ and $\mathbf{x}^{\text{nom}}_{i+1}$ belonging to $\sigma^{\text{nom}}$, are connected by piece-wise linear segments. 
Let $\sigma^{\text{dev}}$ be the computed path deviation from the state $\mathbf{x}_{\text{start}} \in \sigma^{\text{nom}}$ to the end state $\mathbf{x}_{\text{end}} \in \sigma^{\text{nom}}$, i.e. 
\begin{equation}
\sigma^{\text{dev}} = \left(\mathbf{x}^{\text{dev}}_{k}\right)_{k=1}^N
\end{equation}
where $\mathbf{x}^{\text{dev}}_{1} = \mathbf{x}_{\text{start}}$ and $\mathbf{x}^{\text{dev}}_{N} = \mathbf{x}_{\text{end}}$.

The cost function that penalizes deviations from the nominal path is defined as the distance of each state in the path $\sigma^{\text{dev}}$ to the closest point in the nominal path $\sigma^{\text{nom}}$, as follows
\begin{equation}\label{eq:cost_function}
    c_d(\sigma^{\text{dev}}) \triangleq \sum^{N}_{k=1} \min\norm{\sigma^{\text{nom}} - \mathbf{x}^{\text{dev}}_k}_2, \quad \forall \, \mathbf{x}_k \in \sigma^{\text{dev}}
\end{equation}
which yields solutions that tend towards the nominal path. However, depending on the length of each segment in $\sigma^{\text{nom}}$ and the underlying steering function, minimizing the proposed cost function may result in corner cutting behaviour at the transition between two nominal path segments.

\begin{figure}[t]
    % trim={<left> <lower> <right> <upper>}
     \centering
     \begin{subfigure}[b]{0.32\columnwidth}
         \centering
         \includegraphics[width=\textwidth,trim={2cm 0.5cm 3.5cm 1cm},clip]{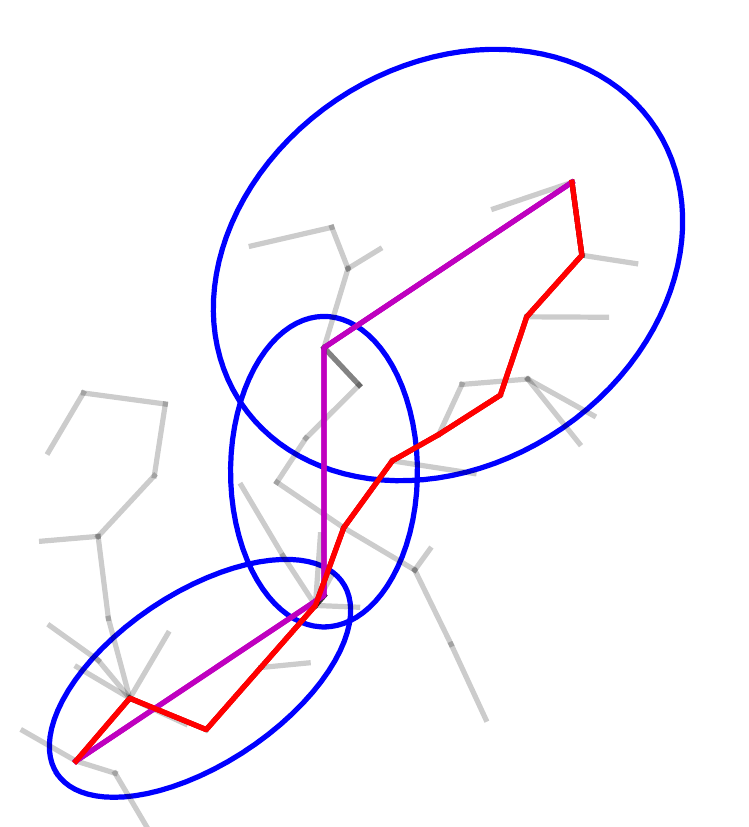}
         \caption{$n = 80$}
         %\label{fig:y equals x}
     \end{subfigure}
     \begin{subfigure}[b]{0.32\columnwidth}
         \centering
         \includegraphics[width=\textwidth,trim={2cm 0.5cm 3.5cm 1cm},clip]{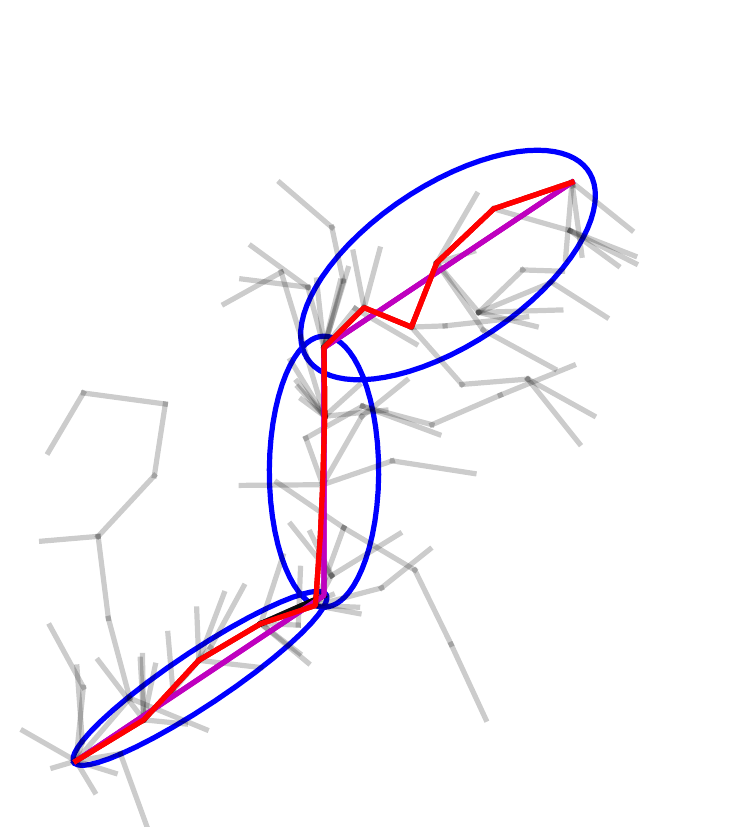}
         \caption{$n=180$}
         %\label{fig:three sin x}
     \end{subfigure}
     \begin{subfigure}[b]{0.32\columnwidth}
         \centering
         \includegraphics[width=\textwidth,trim={2cm 0.5cm 3.5cm 1cm},clip]{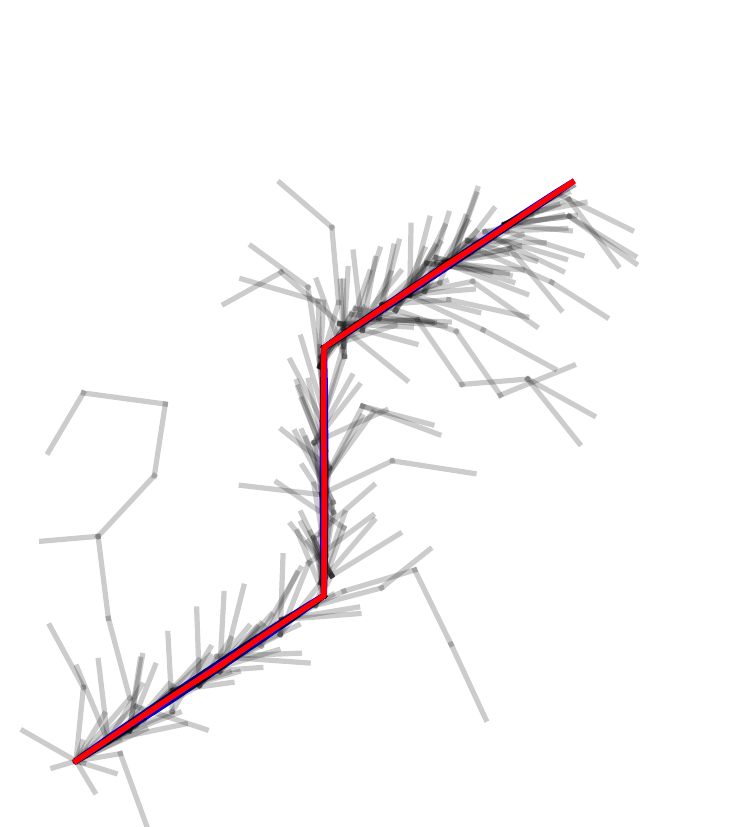}
         \caption{$n = 400$}
         %\label{fig:five over x}
     \end{subfigure}
        \caption{Without obstacles, the informed subsets (blue) computes a path (red) that converges to the nominal (magenta). }
        \label{fig:cost_convergence}
        \vspace{-0.5cm}
\end{figure}

For a tighter fit in the corners, both the nominal and found path can be linearly interpolated, such that the deviation is computed with a resolution $\epsilon$ between each state in the path $\sigma^{\text{dev}}$ towards the interpolated nominal. As the nominal path remains fixed, one can efficiently compute the distance towards it using e.g. a k-d tree. Depending on the tightness required for a given application, one can adjust $\epsilon$ accordingly or entirely skip interpolating. 

The cost of the deviation tends towards the global minimum as the resolution of the nominal and deviation is increased,
\begin{equation}
    \lim_{\epsilon\rightarrow 0} c_d(\sigma^{\text{dev}}) = c_d(\sigma^*)
\end{equation}
where both $\sigma^{\text{dev}}$ and $\sigma^{\text{nom}}$ are linearly interpolated with resolution $\epsilon$. Similarly, for the obstacle free case
\begin{equation}
    \lim_{\epsilon\rightarrow 0} \sigma^{\text{dev}} = \sigma^* = \sigma^{\text{nom}}
\end{equation}
the deviation converges to the global minimum ($\sigma^{\text{nom}})$, which is demonstrated in Fig.~\ref{fig:cost_convergence}.

\begin{rem} The proposed motion planner can be extended to account for multiple objectives, potentially conflicting, by expanding the cost function \eqref{eq:cost_function} with additional terms properly weighted. For instance, if path length should also be in focus, then the following cost function will trade off between path deviation $c_d(\cdot)$ and total path length $c_l(\cdot)$ through the weight $\omega \in [0,1)$
\begin{equation}\label{eq:cost_structure}
    c(\sigma^{\text{dev}}) = (1 - \omega) c_d(\sigma^{\text{dev}}) + \omega c_l(\sigma^{\text{dev}}) .
\end{equation}
\end{rem}

\subsection{Informed sampling for minimizing path deviation}
Given the nominal path $\sigma^{\text{nom}}$ consisting of $m$ states, the proposed informed subset consists of the union of $m-1$ ellipsoids along each nominal path segment, that is
\begin{equation}\label{eq:proposed_union}
    X_{\hat{F}} = \bigcup_{i=1}^{m-1}X_{\hat{f},i}
\end{equation}
where
\begin{align}
    X_{\hat{f},i} = \{ \mathbf{x} \in &\mathcal{X} \, | \nonumber\\ &\norm{\mathbf{x}^{\text{nom}}_{i} - \mathbf{x}}_2 +  \norm{\mathbf{x} - \mathbf{x}^{\text{nom}}_{i+1}}_2 \leq c_{\text{best},i} \} .
\end{align}
When $m=2$ the method defaults to the informed subset from \cite{gammell2014informed}. An important guarantee posed by the informed subset in \cite{gammell2014informed} is that the encompassing ellipsoid guarantees to include all possible points that may improve the current best found solution. It is therefore important that the union of ellipsoids is constructed such that the same guarantee is maintained.

To ensure that the entire path always falls within the joined ellipsoids, the computation of $c_{\text{best},i}$ must share states with the neighbouring ellipsoids. Given the nominal path $\sigma^{\text{nom}}$ there are $m-2$ states $\mathbf{x}^{\text{nom}}_i$ connecting $\mathbf{x}_{\text{start}}$ to $\mathbf{x}_{\text{end}}$ through $\sigma^{\text{nom}}$. Let $\mathcal{N}$ be the finite sequence of common states that are defined as the nearest states in the current path deviation $\sigma^{\text{dev}}$ to each of the $m-2$ nominal states $\mathbf{x}_i^\text{nom}$, i.e.
\begin{align}
    \mathcal{N} &= \left( (\mathbf{x}^*,k)_j\right)_{j = 1}^{m-2}
\end{align}
where
\begin{align}
    \mathbf{x}^* &= \argmin_{\mathbf{x}^{\text{dev}} \in \sigma^{\text{dev}}}  \norm{\mathbf{x}^{\text{dev}} - \mathbf{x}_i^{\text{nom}}}_2, \quad \forall \, i = 2, \ldots , m-1
\end{align}
and $k$ is the index identifying the position of the state $\mathbf{x}^*$ in the path deviation $\sigma^{\text{dev}}$.  

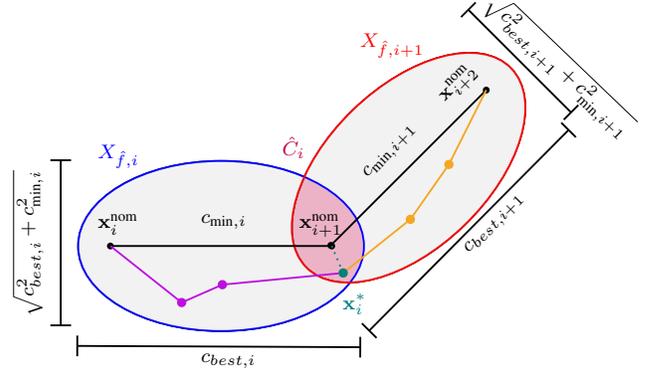
\begin{figure}[t]
    \centering
    \resizebox{\columnwidth}{!}{\tikzset{every picture/.style={line width=0.75pt}} %set default line width to 0.75pt        

\begin{tikzpicture}[x=0.75pt,y=0.75pt,yscale=-1,xscale=1]
%uncomment if require: \path (0,993); %set diagram left start at 0, and has height of 993

%Shape: Ellipse [id:dp4351989286279059] 
\draw[name path=A] (269.04,600.58) .. controls (269.04,575.17) and (303.62,554.56) .. (346.29,554.56) .. controls (388.96,554.56) and (423.55,575.17) .. (423.55,600.58) .. controls (423.55,626) and (388.96,646.6) .. (346.29,646.6) .. controls (303.62,646.6) and (269.04,626) .. (269.04,600.58) -- cycle ;
%Shape: Ellipse [id:dp3089696678586269] 
\draw[name path=B, fill=purple!30] (392.22,612.02) .. controls (375.55,594.84) and (386.85,556.83) .. (417.47,527.11) .. controls (448.09,497.4) and (486.43,487.24) .. (503.1,504.42) .. controls (519.78,521.6) and (508.47,559.61) .. (477.85,589.33) .. controls (447.23,619.04) and (408.89,629.2) .. (392.22,612.02) -- cycle ;
\tikzfillbetween[of=A and B]{lightgray!20};
% Extra ellipses for the borders
\draw[blue] (269.04,600.58) .. controls (269.04,575.17) and (303.62,554.56) .. (346.29,554.56) .. controls (388.96,554.56) and (423.55,575.17) .. (423.55,600.58) .. controls (423.55,626) and (388.96,646.6) .. (346.29,646.6) .. controls (303.62,646.6) and (269.04,626) .. (269.04,600.58) -- cycle ;
\draw[red] (392.22,612.02) .. controls (375.55,594.84) and (386.85,556.83) .. (417.47,527.11) .. controls (448.09,497.4) and (486.43,487.24) .. (503.1,504.42) .. controls (519.78,521.6) and (508.47,559.61) .. (477.85,589.33) .. controls (447.23,619.04) and (408.89,629.2) .. (392.22,612.02) -- cycle ;
%Straight Lines [id:da30087009182829694] 
\draw    (286.46,600.58) -- (406.13,600.58) ;
\draw [shift={(406.13,600.58)}, rotate = 0] [color={rgb, 255:red, 0; green, 0; blue, 0 }  ][fill={rgb, 255:red, 0; green, 0; blue, 0 }  ][line width=0.75]      (0, 0) circle [x radius= 1.34, y radius= 1.34]   ;
\draw [shift={(286.46,600.58)}, rotate = 0] [color={rgb, 255:red, 0; green, 0; blue, 0 }  ][fill={rgb, 255:red, 0; green, 0; blue, 0 }  ][line width=0.75]      (0, 0) circle [x radius= 1.34, y radius= 1.34]   ;

%Straight Lines [id:da8387294805934922] 
\draw    (405.66,600.22) -- (489.67,516.21) ;
\draw [shift={(489.67,516.21)}, rotate = 315] [color={rgb, 255:red, 0; green, 0; blue, 0 }  ][fill={rgb, 255:red, 0; green, 0; blue, 0 }  ][line width=0.75]      (0, 0) circle [x radius= 1.34, y radius= 1.34]   ;
\draw [shift={(405.66,600.22)}, rotate = 315] [color={rgb, 255:red, 0; green, 0; blue, 0 }  ][fill={rgb, 255:red, 0; green, 0; blue, 0 }  ][line width=0.75]      (0, 0) circle [x radius= 1.34, y radius= 1.34]   ;
%Straight Lines [id:da08930534843716864] 
\draw [color={rgb, 255:red, 189; green, 16; blue, 224 }  ,draw opacity=1 ]   (286.46,600.58) -- (325,631.17) ;
\draw [shift={(325,631.17)}, rotate = 38.43] [color={rgb, 255:red, 189; green, 16; blue, 224 }  ,draw opacity=1 ][fill={rgb, 255:red, 189; green, 16; blue, 224 }  ,fill opacity=1 ][line width=0.75]      (0, 0) circle [x radius= 2.01, y radius= 2.01]   ;
%Straight Lines [id:da13494788244474343] 
\draw [color={rgb, 255:red, 189; green, 16; blue, 224 }  ,draw opacity=1 ]   (347,621.5) -- (412.33,615.17) ;
%Straight Lines [id:da7178823368458322] 
\draw [color={rgb, 255:red, 245; green, 166; blue, 35 }  ,draw opacity=1 ]   (412.33,615.17) -- (448.67,586.17) ;
\draw [shift={(448.67,586.17)}, rotate = 321.4] [color={rgb, 255:red, 245; green, 166; blue, 35 }  ,draw opacity=1 ][fill={rgb, 255:red, 245; green, 166; blue, 35 }  ,fill opacity=1 ][line width=0.75]      (0, 0) circle [x radius= 2.01, y radius= 2.01]   ;
\draw [shift={(412.33,615.17)}, rotate = 321.4] [color={rgb, 255:red, 245; green, 166; blue, 35 }  ,draw opacity=1 ][fill={rgb, 255:red, 245; green, 166; blue, 35 }  ,fill opacity=1 ][line width=0.75]      (0, 0) circle [x radius= 2.01, y radius= 2.01]   ;
%Straight Lines [id:da9557160594946945] 
\draw [color={rgb, 255:red, 245; green, 166; blue, 35 }  ,draw opacity=1 ]   (448.67,586.17) -- (469.49,556.44) ;
%Straight Lines [id:da8986634786124554] 
\draw [color={rgb, 255:red, 245; green, 166; blue, 35 }  ,draw opacity=1 ]   (469.49,556.44) -- (489.67,516.21) ;
\draw [shift={(469.49,556.44)}, rotate = 296.64] [color={rgb, 255:red, 245; green, 166; blue, 35 }  ,draw opacity=1 ][fill={rgb, 255:red, 245; green, 166; blue, 35 }  ,fill opacity=1 ][line width=0.75]      (0, 0) circle [x radius= 2.01, y radius= 2.01]   ;
%Straight Lines [id:da19277550466123206] 
\draw [dotted][color=teal ,draw opacity=1 ]   (406.13,600.58) -- (412.33,615.17) ;
\draw [shift={(412.33,615.17)}, rotate = 336.28] [color=teal  ,draw opacity=1 ][fill=teal  ,fill opacity=1 ][line width=0.75]      (0, 0) circle [x radius= 2.01, y radius= 2.01]   ;
%Straight Lines [id:da7928916354101898] 
\draw [color={rgb, 255:red, 189; green, 16; blue, 224 }  ,draw opacity=1 ]   (325,631.17) -- (347,621.5) ;
\draw [shift={(347,621.5)}, rotate = 336.28] [color={rgb, 255:red, 189; green, 16; blue, 224 }  ,draw opacity=1 ][fill={rgb, 255:red, 189; green, 16; blue, 224 }  ,fill opacity=1 ][line width=0.75]      (0, 0) circle [x radius= 2.01, y radius= 2.01]   ;
%Straight Lines [id:da9894077810075115] 
\draw    (259.61,643.5) -- (259.61,554.17) ;
\draw [shift={(259.61,554.17)}, rotate = 90] [color={rgb, 255:red, 0; green, 0; blue, 0 }  ][line width=0.75]    (0,5.59) -- (0,-5.59)   ;
\draw [shift={(259.61,643.5)}, rotate = 90] [color={rgb, 255:red, 0; green, 0; blue, 0 }  ][line width=0.75]    (0,5.59) -- (0,-5.59)   ;
%Straight Lines [id:da08953465584002318] 
\draw    (268.56,654.64) -- (421.56,655.33) ;
\draw [shift={(421.56,655.33)}, rotate = 180.26] [color={rgb, 255:red, 0; green, 0; blue, 0 }  ][line width=0.75]    (0,5.59) -- (0,-5.59)   ;
\draw [shift={(268.56,654.64)}, rotate = 180.26] [color={rgb, 255:red, 0; green, 0; blue, 0 }  ][line width=0.75]    (0,5.59) -- (0,-5.59)   ;
%Straight Lines [id:da7189852388936886] 
\draw    (535.53,528.28) -- (480,472.75) ;
\draw [shift={(480,472.75)}, rotate = 45] [color={rgb, 255:red, 0; green, 0; blue, 0 }  ][line width=0.75]    (0,5.59) -- (0,-5.59)   ;
\draw [shift={(535.53,528.28)}, rotate = 45] [color={rgb, 255:red, 0; green, 0; blue, 0 }  ][line width=0.75]    (0,5.59) -- (0,-5.59)   ;
%Straight Lines [id:da7276147700039641] 
\draw    (426.41,646.38) -- (534.2,537.8) ;
\draw [shift={(534.2,537.8)}, rotate = 134.79] [color={rgb, 255:red, 0; green, 0; blue, 0 }  ][line width=0.75]    (0,5.59) -- (0,-5.59)   ;
\draw [shift={(426.41,646.38)}, rotate = 134.79] [color={rgb, 255:red, 0; green, 0; blue, 0 }  ][line width=0.75]    (0,5.59) -- (0,-5.59)   ;

% Text Node
\draw (232,640) node [anchor=north west][inner sep=0.75pt]  [font=\footnotesize,rotate=-270] [align=left] {$\displaystyle \sqrt{c_{best,i}^{2} + c_{\text{min},i}^{2}}$};
% Text Node
\draw (334,657.14) node [anchor=north west][inner sep=0.75pt]  [font=\footnotesize] [align=left] {$\displaystyle c_{best,i}$};
% Text Node
\draw (497.14,459.2) node [anchor=north west][inner sep=0.75pt]  [font=\footnotesize,rotate=-43.79] [align=left] {$\displaystyle \sqrt{c_{best,i+1}^{2} + c_{\text{min},i+1}^{2}}$};
% Text Node
\draw (474.7,600.86) node [anchor=north west][inner sep=0.75pt]  [font=\footnotesize,rotate=-314.53] [align=left] {$\displaystyle c_{best,i+1}$};
% Text Node
\draw (378,540) node [anchor=north west][inner sep=0.75pt]  [font=\footnotesize][text=purple] [align=left] {$\hat{C}_i$};
% Text Node
\draw (278.2,544.19) node [anchor=north west][inner sep=0.75pt]  [font=\footnotesize, blue] [align=left] {$X_{\hat{f},i}$};
% Text Node
\draw (420,484.19) node [anchor=north west][inner sep=0.75pt]  [font=\footnotesize, red] [align=left] {$X_{\hat{f},i+1}$};
% Text Node
\draw (278.2,582) node [anchor=north west][inner sep=0.75pt]  [font=\footnotesize] [align=left] {$\mathbf{x}_{i}^{{\text{nom}}}$};
% Text Node
\draw (387,582) node [anchor=north west][inner sep=0.75pt]  [font=\footnotesize] [align=left] {$\mathbf{x}_{i+1}^{{\text{nom}}}$};
% Text Node
\draw (460,517) node [anchor=north west][inner sep=0.75pt]  [font=\footnotesize,rotate=-314.53] [align=left] {$\mathbf{x}_{i+2}^{{\text{nom}}}$};
% Text Node
\draw (334,582) node [anchor=north west][inner sep=0.75pt]  [font=\footnotesize] [align=left] {$c_{\text{min},i}$};
% Text Node
\draw (420,560) node [anchor=north west][inner sep=0.75pt]  [font=\footnotesize,rotate=-314.53] [align=left] {$c_{\text{min},i+1}$};
% Text Node
\draw (410.6,625) node [anchor=north west][inner sep=0.75pt]  [font=\footnotesize, teal] [align=left] {$\displaystyle \mathbf{x}_{i}^*$};

\end{tikzpicture}}
    \caption{The proposed informed subset for minimizing path deviations, given a nominal path consisting of $m$ states and $m-1$ piece-wise linear segments. The informed subset is composed of the union of $m-1$ ellipsoids.}
    \label{fig:nleg_informedset}
    \vspace{-0.5cm}
\end{figure}

The corresponding $c_{\text{best},i}$ for each ellipsoid is then computed for $m > 2$,
\begin{equation}
    \mathcal{C}_{best} = \left( c_{\text{best},i} = c_l(\rho_i) \quad \forall \, i = 1, \ldots , m-1\right)
\end{equation}
where
\begin{equation}
    \rho_i = \begin{cases}
    \left(\mathbf{x}_{\text{start}},\mathbf{x}^{\text{dev}}_2,\dots,\mathbf{x}^*_i,\mathbf{x}_{i+1}^{\text{nom}}\right)& \text{if } i = 1\\
    \begin{aligned}[t]\arraycolsep=0pt
    \bigl(&\mathbf{x}_i^{\text{nom}},\mathbf{x}^*_{i-1},\\
    &\quad\mathbf{x}^{\text{dev}}_{k_{i-1}+1},\dots,\mathbf{x}^*_{i},\mathbf{x}_{i+1}^{\text{nom}}\bigr)
    \end{aligned}& \text{if } 1 < i < m - 1\\
    \begin{aligned}[t]\arraycolsep=0pt
    \bigl(&\mathbf{x}_i^{\text{nom}},\mathbf{x}^*_{i-1},\\
    &\quad\mathbf{x}^{\text{dev}}_{k_{i-1}+1},\dots,\mathbf{x}^{\text{dev}}_{N-1},\mathbf{x}_{\text{end}}\bigr)
    \end{aligned}& \text{if } i = m - 1
    \end{cases}
\end{equation}
is the piece-wise continuous part of the current path deviation $\sigma^{\text{dev}}$ contained within an ellipsoid, and connected with the closest corresponding state along the nominal trajectory, as shown in Fig.~\ref{fig:nleg_informedset}.  

Given $X_{\hat{F}}$ and $\mathcal{C}_{\text{best}}$, one can guarantee, by construction, that the current deviation $\sigma^{\text{dev}}$ and all points capable of improving said deviation, are contained within $X_{\hat{F}}$. As a given $\rho_i$ has a state in common with each neighbouring ellipsoid through the node $\mathbf{x}_i^*$, therefore the combined path $\sigma^\text{dev}$ is also guaranteed to exist within the union of ellipsoids.
The proposed subset maintains this property as the deviation converges to the minimum.

Once the sequence of ellipsoids has been constructed, one can sample them using the technique described by \cite{gammell2018informed}, where a given ellipsoid is selected and subsequently uniformly sampled based on its relative measure. Samples are rejected in proportion to their membership of a given ellipsoid, in order maintain uniformity.

\subsection{Sample biasing}\label{sec:sample_biasing}
Sample biasing is a very common technique for improving the performance of SBMP algorithms \cite{veras2019systematic,gammell2021asymptotically,urmson2003approaches}. Most implementations utilize a goal biasing strategy, in order to ensure that $\mathbf{x}_{\text{start}}$ and $\mathbf{x}_{\text{end}}$ connect \cite{LaValle2001}. The bias is introduced by checking if the parameter $0 \leq \delta \leq 1$ is smaller than the uniformly distributed random variable $u \sim \mathcal{U}(0,1)$, and choosing either the goal state or a sample from the space $\mathcal{U}(X_{\text{rect}})$ as the random sample.
This idea is extended to sampling all (except $\mathbf{x}_{\text{start}}$) states belonging to the nominal path $\sigma^{\text{nom}}$, such that
\begin{equation}
    \mathbf{x}_{\text{rand}} = \begin{cases}
    \,\mathcal{U}\left(X_{\text{space}}\right), & \text{if } \delta < u\\
    \,\mathcal{U}\left(\left(\mathbf{x}^{\text{nom}}_1, \dots,\mathbf{x}^{\text{nom}}_m\right)\right), & \text{otherwise }
    \end{cases}
\end{equation}
where $X_{\text{space}}$ is the current sampling space (e.g. $X_{\text{rect}}$ or $X_{\hat{F}}$).

\subsection{Switching condition}
Given certain circumstances, sampling the informed set may be disadvantageous, compared to simply sampling the original space, since the informed set is generated based on the current best solution cost \cite{mandalika2021guided,enevoldsen2021colregs}. With a high cost, the volume of the informed set may be larger than that of the original space. It is therefore natural, and also important, to compute a switching condition, which will determine whether or not the informed set provides sufficient value. 

As with \eqref{eq:nom_switching}, one can compare the Lebesgue measure of $X_{\text{rect}}$ and the proposed informed subset
\begin{equation}
    \lambda \left(X_{\hat{F}} \right) < \lambda \left(X_{\text{rect}} \right)
\end{equation}
with the measure of \eqref{eq:proposed_union} given by
\begin{equation}
    \lambda \left(X_{\hat{F}} \right) = \sum_{i=1}^{m-1} \lambda\left(X_{\hat{f},i}\right) - \sum^{m-2}_{i=1} \lambda(C_i)
\end{equation}
where $C_i = X_{\hat{f},i} \cap X_{\hat{f},i+1}$.  
However, computing the exact intersection measure, especially for higher dimensions, is non-trivial. Instead an estimate of the intersection $\hat{C}_i$ is used.
\begin{equation}\label{eq:leb_est}
    \lambda \left(\hat{X}_{\hat{F}} \right) = \sum_{i=1}^{m-1} \lambda \left(X_{\hat{f},i} \right) - \sum^{m-2}_{i=1} \lambda \left(\hat{C}_i \right)
\end{equation}
One of the simplest estimates is simply setting $\hat{C}_i = 0$, with the result that the estimated measure of the informed set contains twice as much intersection volume. For small $X_{\hat{f},i}$ compared to $X_{\text{rect}}$ the over-representation of the intersections play a small role in the switching condition. However, if one wants to leverage the informed subset as soon as possible, a better estimate of $\hat{C}_i$ is required. 
If one disregards the required computational time, estimating $\hat{C}_i$ can be achieved using a Monte Carlo or hypervoxel based method. The main issue is adequately selecting the number of random samples or size of the hypervoxels, and thereby making a trade off between accuracy and computational effort. 

\begin{rem}
If the proposed informed sampling scheme remains inactive, either due to a poor choice of heuristic or due to restrictions posed by the problem at hand, the default option is to simply uniformly sample $X_{\text{rect}}$, which results in planning performance equal to the underlying SBMP algorithm.
\end{rem}
\begin{figure*}
     \centering
     \begin{subfigure}[b]{0.67\columnwidth}
         \centering
         \includegraphics[width=\textwidth]{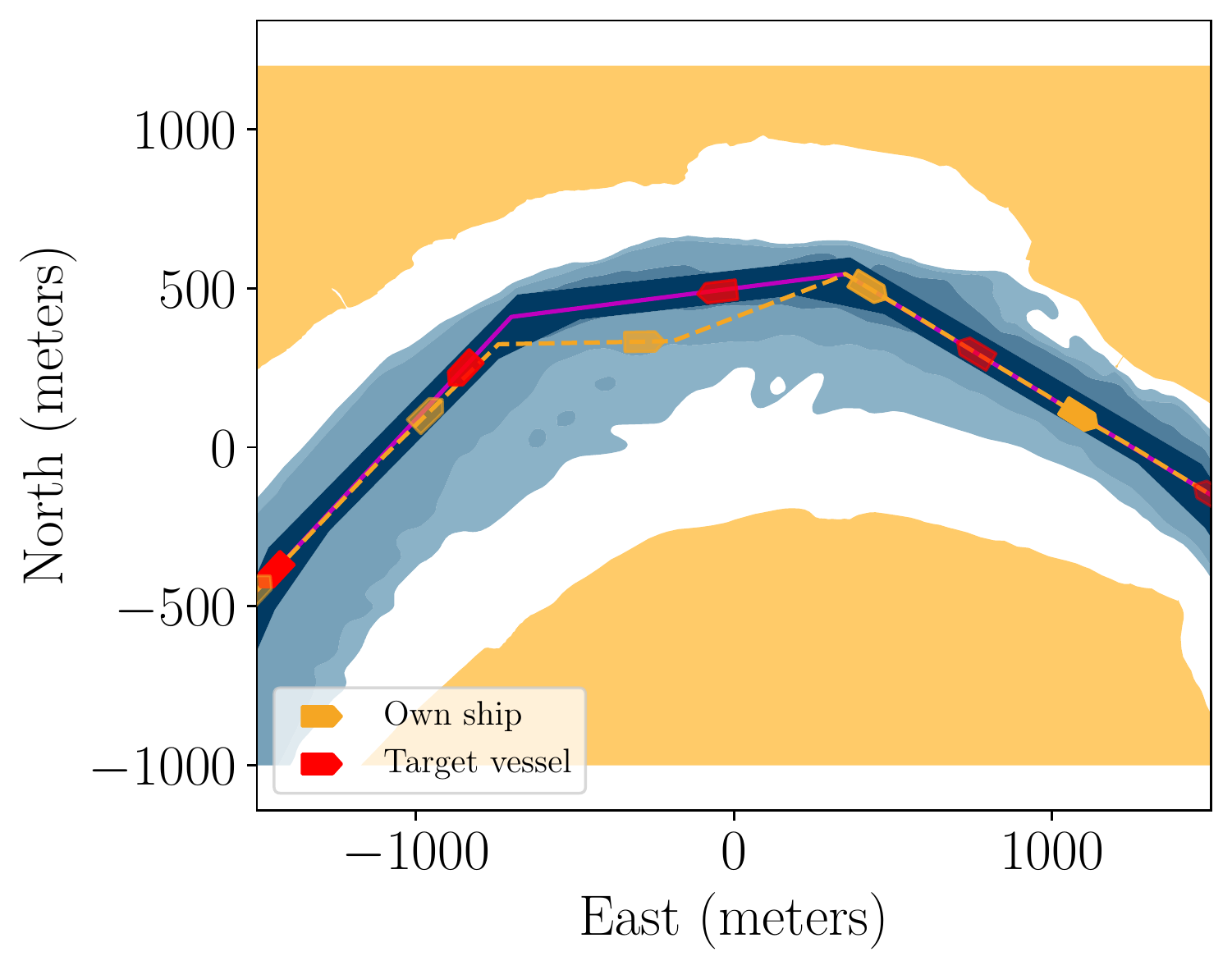}
         \caption{Narrow dredged passage}
         \label{fig:narrow_passage}
     \end{subfigure}
     \begin{subfigure}[b]{0.67\columnwidth}
         \centering
         \includegraphics[width=\textwidth]{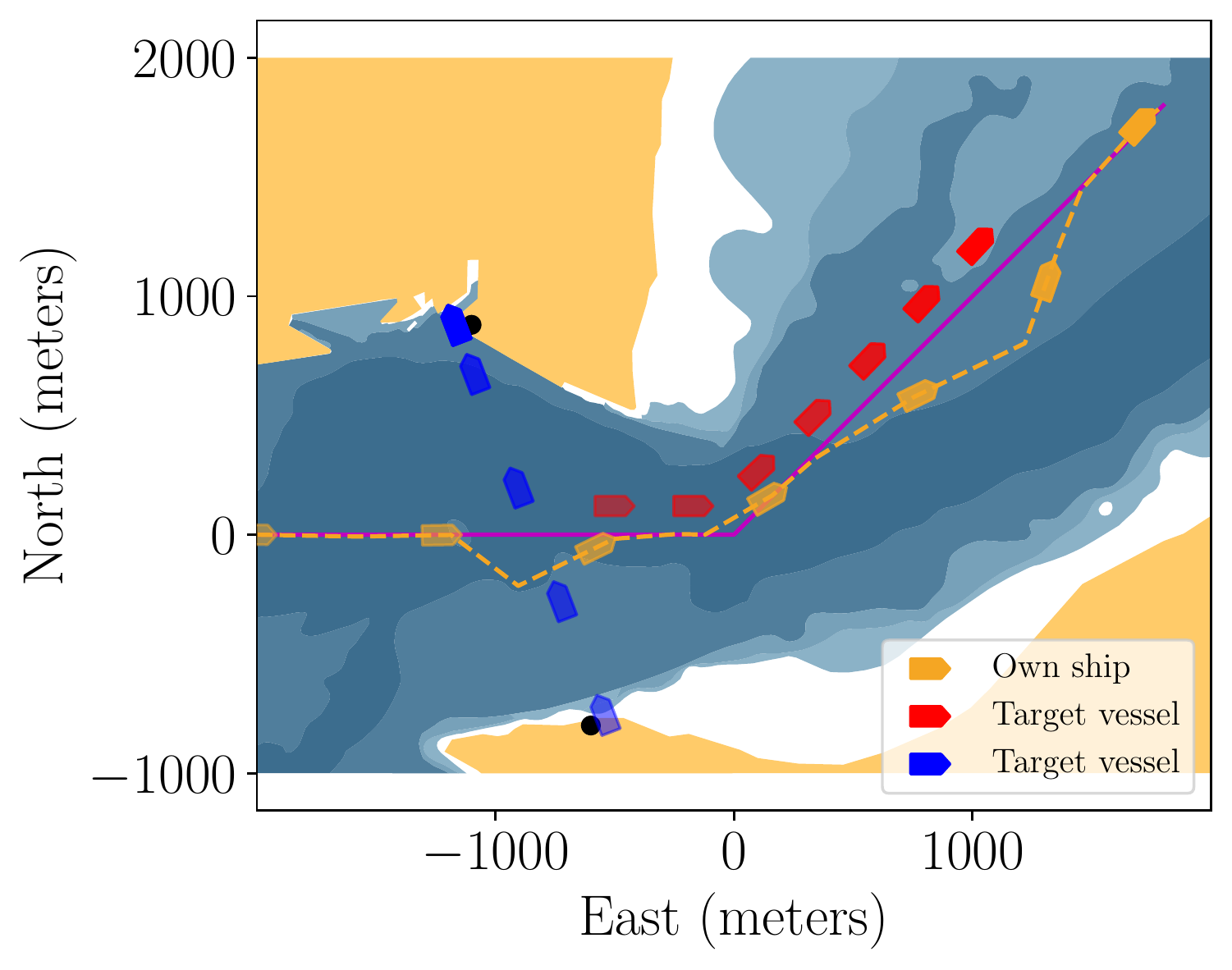}
         \caption{Inner coastal waters}
         \label{fig:inner_coastal}
     \end{subfigure}
     \begin{subfigure}[b]{0.67\columnwidth}
         \centering
         \includegraphics[width=\textwidth]{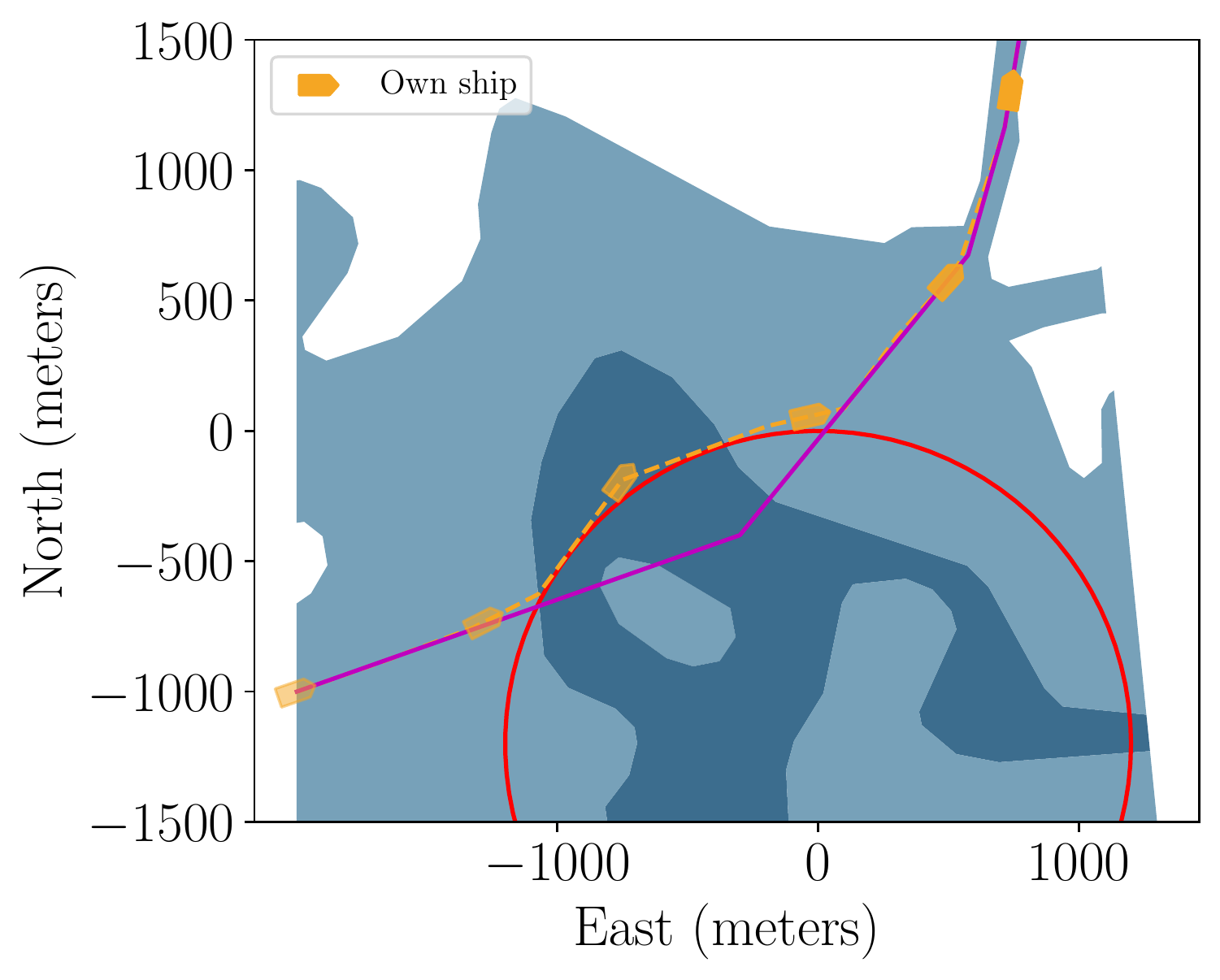}
         \caption{Fjord navigation}
         \label{fig:fishing}
    \end{subfigure}
        \caption{Three collision avoidance scenarios for an autonomous marine vessel, where the nominal path consists of multiple segments. The feasible water depths for the own ship are indicated by the blue polygons, whereas white areas are shallow waters for said ship. The nominal trajectory is shown in magenta. Own ship is marked by the yellow markers and trajectory.}
        \label{fig:vessel_example}
        % \vspace{-0.5cm}
\end{figure*}

\begin{figure*}
     \centering
     \begin{subfigure}[b]{0.67\columnwidth}
         \centering
         \includegraphics[width=\textwidth]{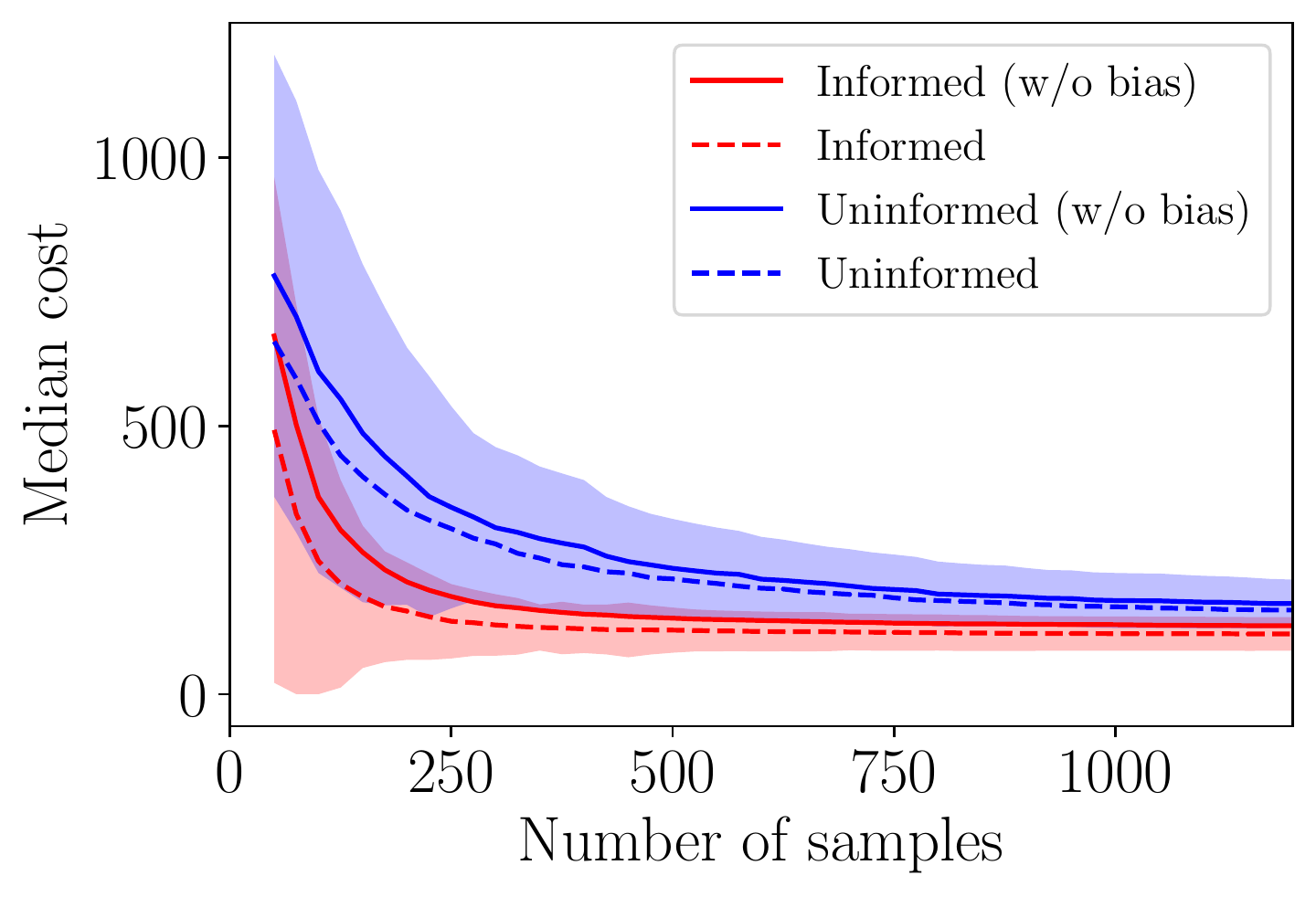}
         \caption{Narrow dredged passage}
         \label{fig:perf_narrow_passage}
     \end{subfigure}
     \begin{subfigure}[b]{0.67\columnwidth}
         \centering
         \includegraphics[width=\textwidth]{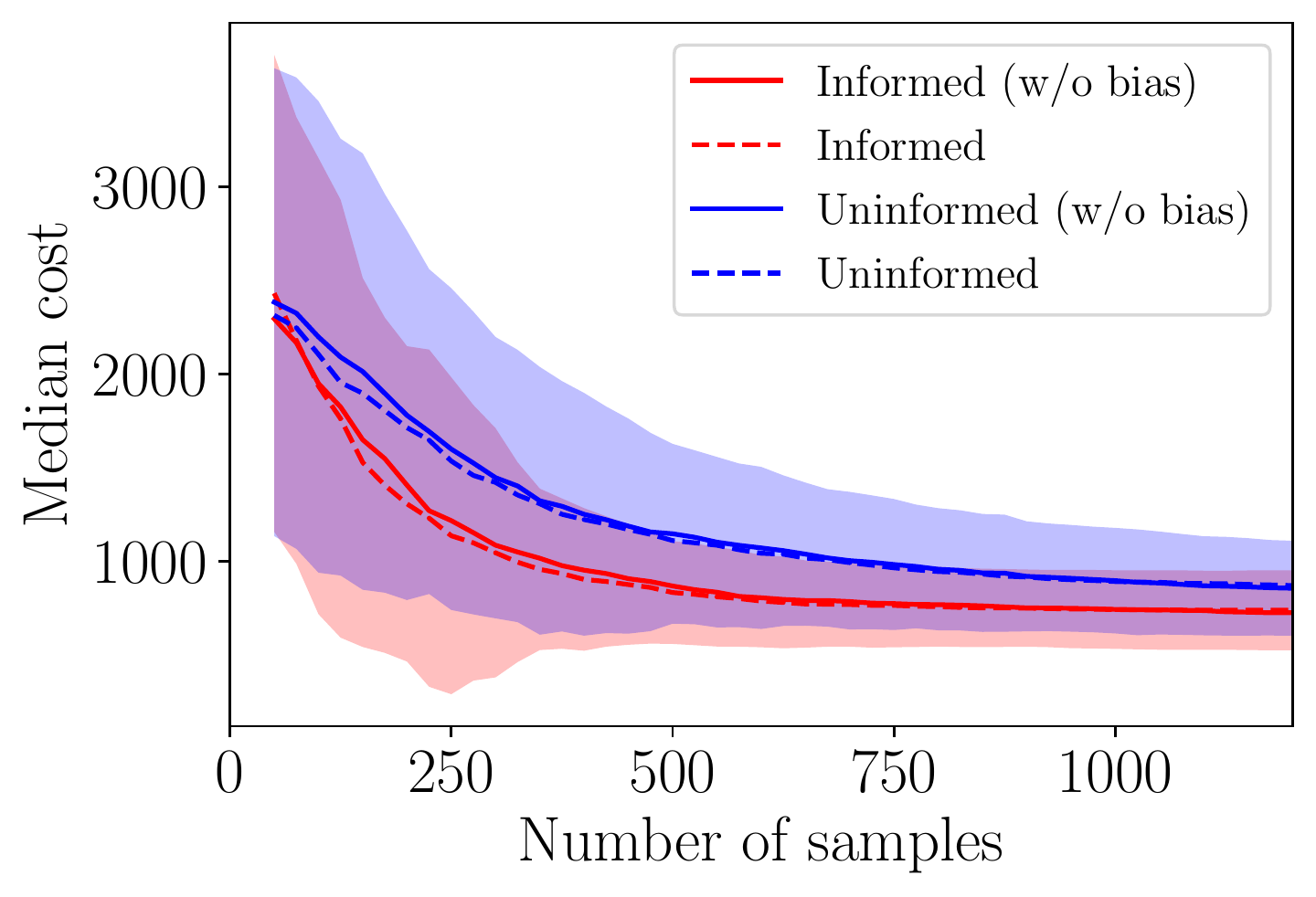}
         \caption{Inner coastal waters}
         \label{fig:perf_inner_coastal}
     \end{subfigure}
     \begin{subfigure}[b]{0.67\columnwidth}
         \centering
         \includegraphics[width=\textwidth]{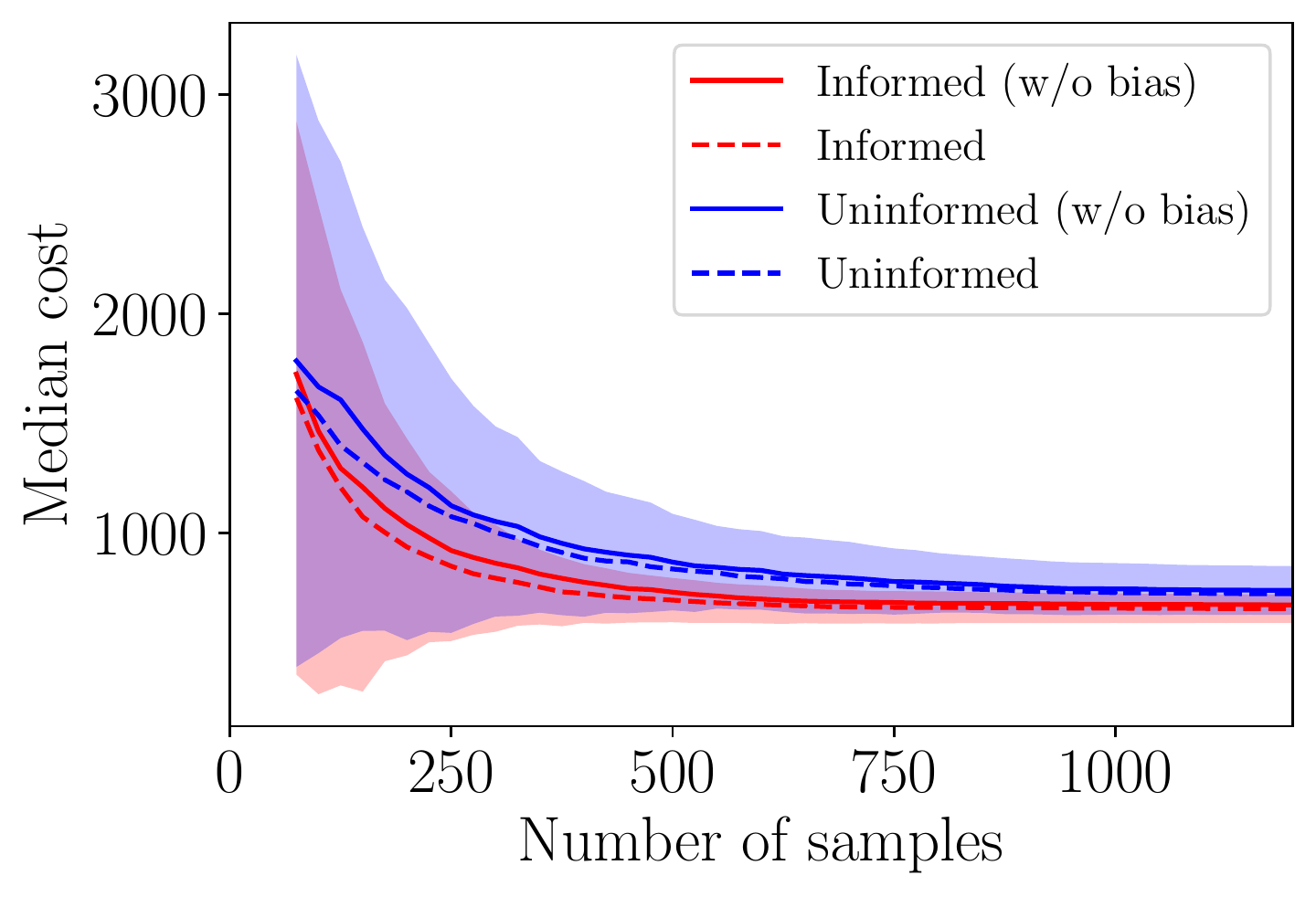}
         \caption{Fjord navigation}
         \label{fig:perf_fishing}
     \end{subfigure}
        \caption{Median costs (250 runs) for each of the scenarios for the autonomous marine craft. The shaded region is the non-parametric $95\%$ confidence interval, and is computed based on the informed and uninformed (with no bias) schemes. The proposed informed scheme converges faster than the uninformed, and is capable of achieving an overall lower solution cost.}
        \label{fig:performance_plots}
        %\vspace{-0.25cm}
\end{figure*}

\begin{figure*}
     \centering
     \begin{subfigure}[b]{0.67\columnwidth}
         \centering
         \includegraphics[width=\textwidth]{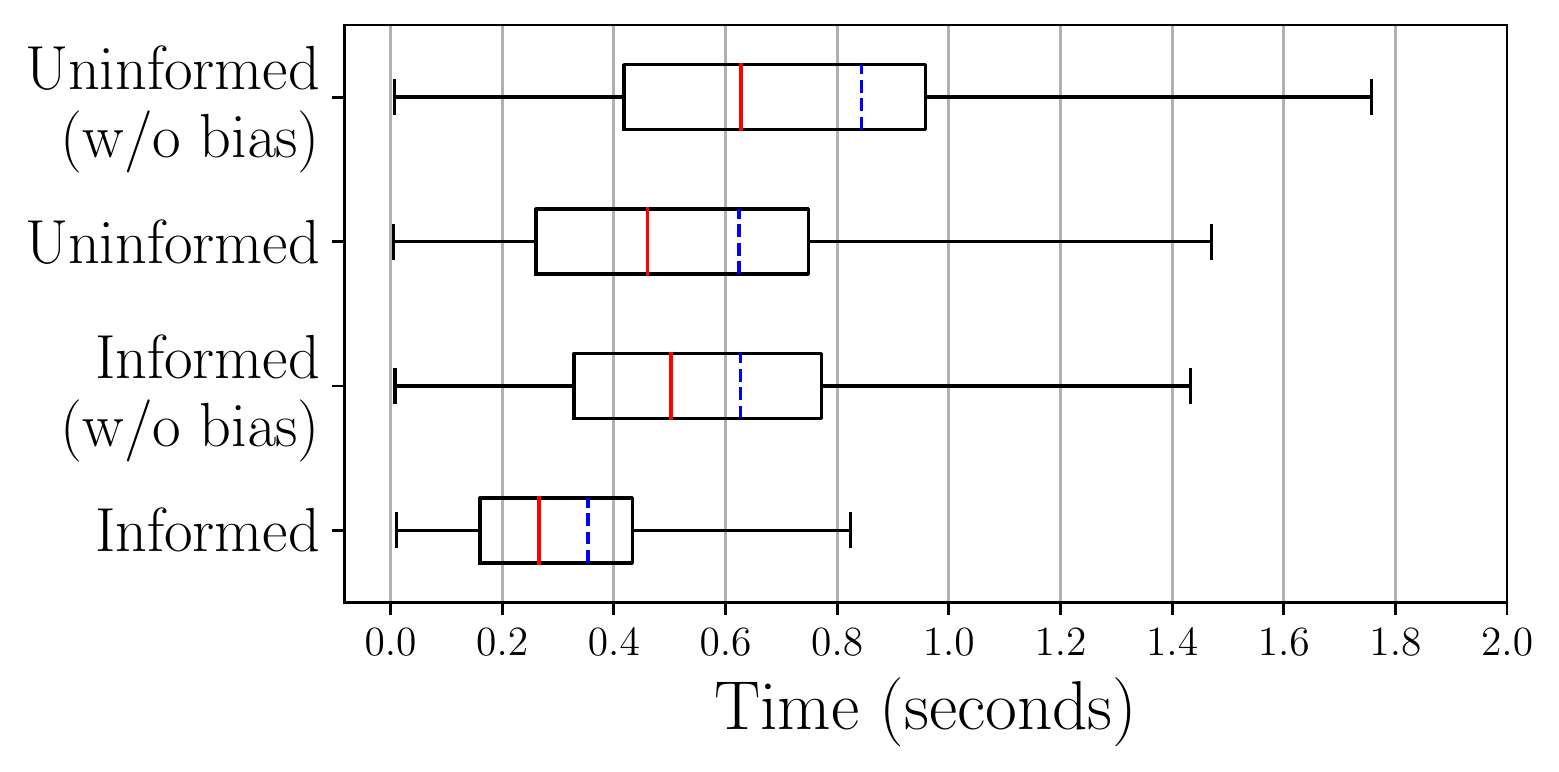}
         \caption{Narrow dredged passage}
         \label{fig:box_narrow_passage}
     \end{subfigure}
     \begin{subfigure}[b]{0.67\columnwidth}
         \centering
         \includegraphics[width=\textwidth]{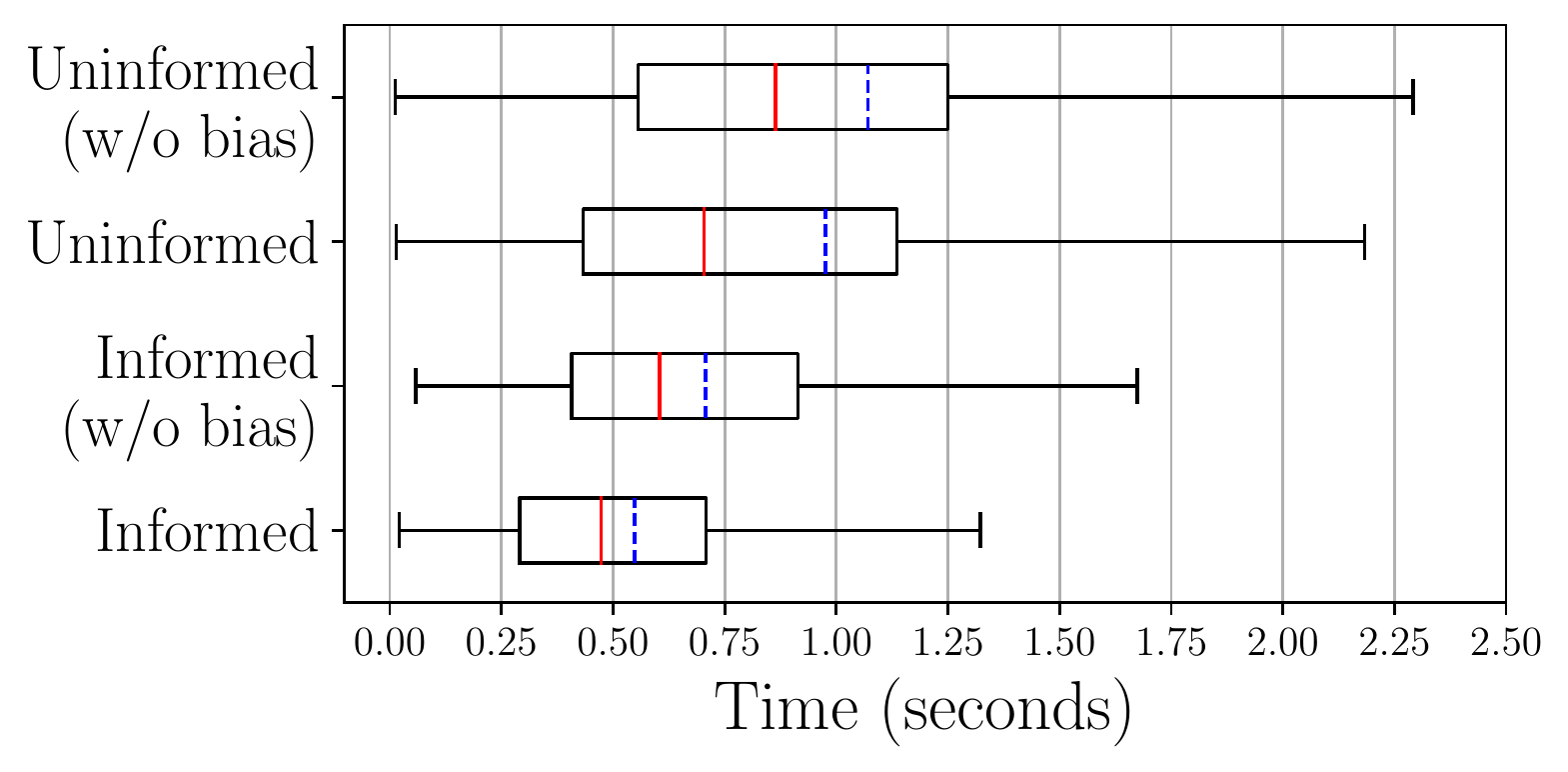}
         \caption{Inner coastal waters}
         \label{fig:box_inner_coastal}
     \end{subfigure}
     \begin{subfigure}[b]{0.67\columnwidth}
         \centering
         \includegraphics[width=\textwidth]{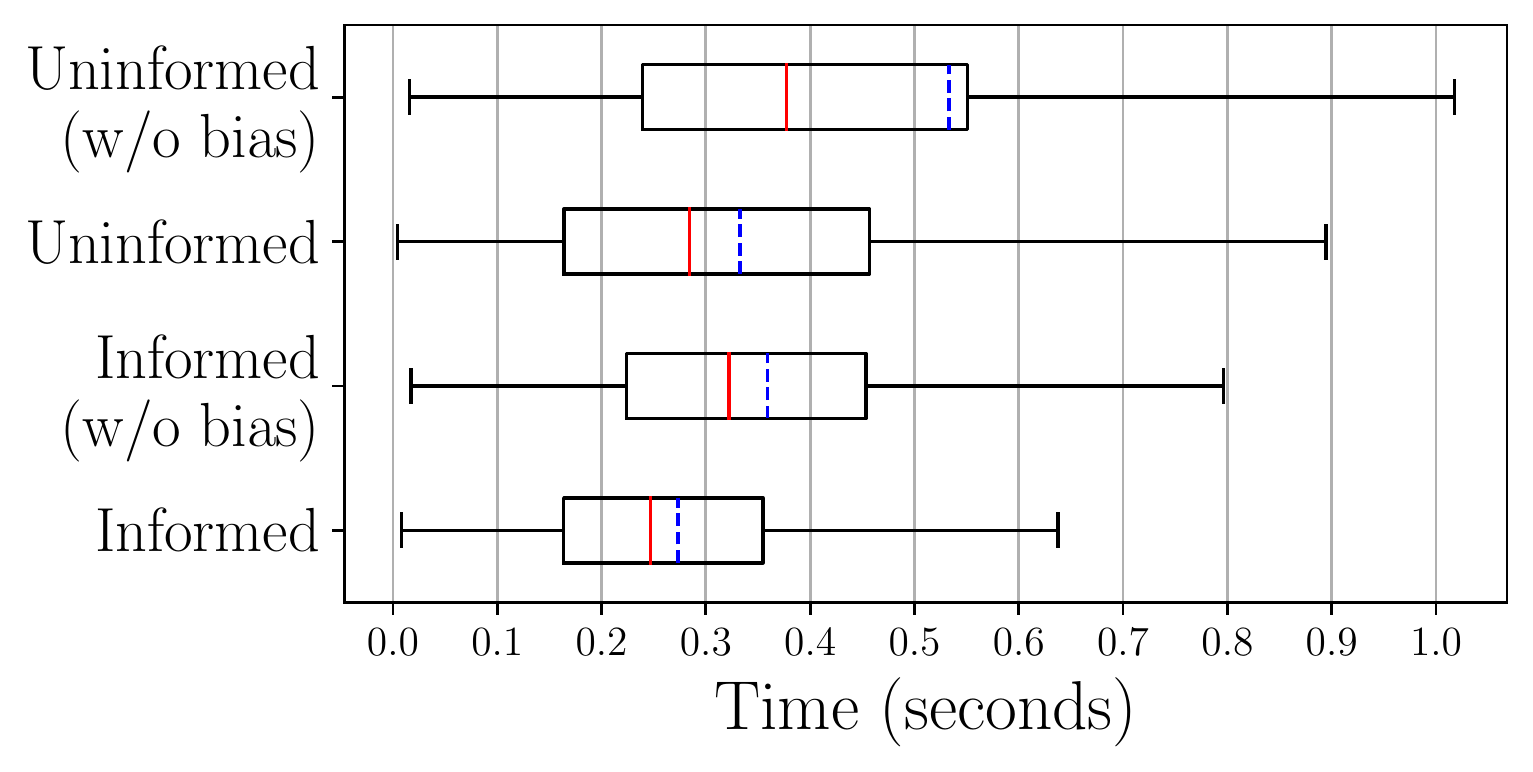}
         \caption{Fjord navigation}
         \label{fig:box_fishing}
     \end{subfigure}
        \caption{Time required to reach $3\sigma$ of the cost, where $\sigma$ is the non-parametric standard deviation computed based on the experiments in Fig.~\ref{fig:performance_plots}. The computational time for each of the four algorithmic variations is evaluated over 1000 trials on the three specified scenarios. The proposed informed method outperforms the uninformed counterpart for all three scenarios, performing a factor of $1.8$ (\ref{fig:box_narrow_passage}), $2.3$ (\ref{fig:box_inner_coastal}) and $1.5$ (\ref{fig:box_fishing}) times better, when comparing median computational times.}
        \label{fig:box_plots}
        \vspace{-0.4cm}
\end{figure*}

\section{Results and Discussion}
The proposed method is tested on three different planning scenarios to achieve collision avoidance of an autonomous surface vessel. 

For autonomous marine crafts, the nominal route is computed prior to vessel departure according to some specifications and is optimized with respect to many important criteria, such arrival time, safety, weather, grounding risk, etc. In the event of potential collision with other vessels, the planner should compute a path deviation that achieves safe and compliant navigation of own ship. However, it is desired that the path alteration remains as close to the nominal path as possible, to ensure the minimum impact on the overall journey performance parameters (arrival time, fuel consumption, etc.), and to avoid endangering the vessel if navigates in coastal waters (see \cite{Enevoldsen2022} and \cite{enevoldsen2021grounding}).

Each simulation case study uses the same baseline SBMP algorithm, RRT*, with its basic parameters unchanged throughout all experiments. The baseline algorithm is referred to as the uninformed method.
This section contains a comparison study between the proposed informed scheme and the uninformed one, with and without the sampling bias. Each simulation assumes $\hat{C}_i=0$ for the estimate of the Lebesgue measure of $\hat{X}_{\hat{F}}$ and the cost function \eqref{eq:cost_function} to achieve the least path deviation. The target vessels are moving along piecewise linear trajectories with arbitrary but known constant speed and heading, as provided, e.g., by the radar. Furthermore, the static obstacles in the environment are represented as polygons and circles, and the target vessels as moving elliptical obstacles. Lastly, the maritime rules-of-the-road (COLREGs rules 13-15) and maneuvering restrictions are enforced as in \cite{Enevoldsen2022}.

\subsection{Case study I - Narrow passages} In confined waters, the navigation within narrow passages is a common occurrence. Here, vessels can be severely constrained by the environment and have limited room to manoeuvre with respect to one another. Therefore the nominal path is planned to minimize the grounding risk. Fig.~\ref{fig:narrow_passage} shows a potential head-on collision scenario, where own ship must deviate. Due to the narrow passage, remaining as close to the nominal as possible is highly important. The proposed method is capable of computing a path that avoids collision while tightly following the nominal path.
\subsection{Case study II - Inner coastal waters} As one leaves the open seas and enters the inner coastal waters, both the traffic scenarios and environmental constraints may change drastically. Here, new obstacles such as active ferries, stationary vessels, etc., must be avoided, all while ensuring a safe distance is kept from shallow waters. Fig.~\ref{fig:inner_coastal} details a scenario in which own ship is travelling through inner coastal waters and must yield for a starboard crossing ferry, as well as overtake a slower vessel. It is demonstrated that the proposed method is able to minimize multiple deviations from the nominal path, such that both collision scenarios are dealt with, whilst maintaining a minimal path deviation.
\subsection{Case study III - Fjord navigation} In some circumstances, the traversable area changes very rapidly, such as the scenario within a fjord depicted in Fig.~\ref{fig:fishing}. Here own ship is trying to leave the fjord, when its nominal path is obstructed by a fishing boat in action. Since the exit of the fjord is so narrow, it is crucial that the planned deviation remains close and converges to the nominal path. The planned deviation successfully avoids the active fishing region and safely converges to the nominal path.

\subsection{Analysis and observations}
The three autonomous ship scenarios demonstrate the proposed schemes ability to effectively compute paths that minimize the deviation from the nominal. Fig.~\ref{fig:performance_plots} demonstrates the informed sets ability to converge to the minimum, and in general an overall lower cost, within a shorter amount of samples compared to the uninformed solution. The results also highlights the impact of the proposed sampling bias, where the bias accelerates the convergence for the informed case. Importantly, a comparison where both the informed and uninformed scheme utilizes the bias was carried out, in order to show that the informed subset is the main contributor to the convergence rate.
Fig.~\ref{fig:box_plots} details the computational times for each of the three proposed scenarios. The informed scheme is able to obtain solutions at a greater rate, despite the additional computational complexity of the proposed informed sampling routine. Overall, the proposed scheme generates solutions at $1.5$-$2.3$ times the rate of the uninformed method while also consistently having the smallest standard deviation, although a suboptimal heuristic for the Lebesgue measure (i.e. $\hat{C}_i = 0$) is used. It is also worth noting that the difference in performance decreases as the area ratio $A_r = \lambda(X_{\mathrm{free,static}})/\lambda(X_{\mathrm{space}})$ increases, where $X_{\mathrm{free,static}}$ is the free space accounting only for the static obstacles. This is reflected by the increased overlap of the confidence intervals for Case study II ($A_r = 63.3\%$, Figs.~\ref{fig:perf_inner_coastal}-\ref{fig:box_inner_coastal}) and Case study III ($A_r = 80.2\%$, Figs.~\ref{fig:perf_fishing}-\ref{fig:box_fishing}). For comparison the area ratio of Case study I is $A_r = 26.5\%$.

As the minimal deviation converges to the nominal, the overall path length may increase, compared to simply minimizing the path length. This can be observed in Fig.~\ref{fig:teaser_figure}, where the informed set initially decreases in volume (Fig.~\ref{fig:teaser_n_250}-\ref{fig:teaser_n_750}) as the path improves towards the nominal, however as the path fully converges to the minimum cost (Fig.~\ref{fig:teaser_n_1000}), the volume of the ellipsoids increase. 
This is due to the construction of $X_{\hat{F}}$, since each ellipsoid is scaled based on the ``local'' path length with respect to a given nominal segment. As the path finds a tighter fit around the obstacles (minimizing the deviation), the overall path length increases.
However, despite the increase or decrease in volume, the informed subset still guarantees that no solution that may improve the current best found cost is omitted. 

\section{Conclusions}
In this paper, the collision avoidance for $n$-dimensional systems having an (optimal) nominal path is addressed by introducing a cost function and informed sampling space for computing solutions with minimum deviation from such a nominal path. Furthermore, the need for a heuristic to estimate the volume spanned by the subset is discussed, with the paper proposing a computationally cheap metric, at the price of a conservative switching condition.
The extension to the informed subset allows the scheme to focus its sampling effort in the neighbourhood surrounding the nominal path, resulting in an accelerated convergence to paths that minimally deviate from the nominal. 
The proposed method is demonstrated on three case studies related to an autonomous marine craft, where the simulated scenarios showed that the proposed method effectively converges to the minimum deviation, at a rate faster than the baseline uninformed method, with the sampling bias further improving the convergence rate of both the informed and uninformed methods. This performance increase is obtained despite using a suboptimal switching condition in the form of the conservative estimate of the Lebesgue measure.

%--------- End of paper
%\addtolength{\textheight}{-8cm}   % This command serves to balance the column lengths

%%%%%%%%%%%%%%%%%%%%%%%%%%%%%%%%%%%%%%%%%%%%%%%%%%%%%%%%%%%%%%%%%%%%%%%%%%%%%%%%

%%%%%%%%%%%%%%%%%%%%%%%%%%%%%%%%%%%%%%%%%%%%%%%%%%%%%%%%%%%%%%%%%%%%%%%%%%%%%%%%

%%%%%%%%%%%%%%%%%%%%%%%%%%%%%%%%%%%%%%%%%%%%%%%%%%%%%%%%%%%%%%%%%%%%%%%%%%%%%%%%
%\section*{APPENDIX}

%\section*{ACKNOWLEDGMENT}
%\hl{Acknowledgment}
%%%%%%%%%%%%%%%%%%%%%%%%%%%%%%%%%%%%%%%%%%%%%%%%%%%%%%%%%%%%%%%%%%%%%%%%%%%%%%%%

\bibliographystyle{IEEEtran}
\bibliography{mainbib.bib}

\end{document}